\documentclass{elsarticle}

\usepackage{graphicx}
\usepackage{subfigure}
\usepackage{algorithm}
\usepackage[noend]{algorithmic} 
\usepackage{amsfonts, amssymb, amsmath}
\usepackage[noend]{algorithmic} 
\usepackage[english]{babel}
\usepackage{slashbox}

\newtheorem{thm}{Theorem}
\newtheorem{lem}[thm]{Lemma}

\def\sc{{\mathbf sc}}
\def\s{{\mathbf s}}

\def\h{{\mathbf h}}
\def\w{{\mathbf w}}

\def\d{{\mathbf d}}
\def\g{{\mathbf g}}
\def\e{{\mathbf e}}
\def\m{{\mathbf m}}
\def\f{{\mathbf f}}

\def\G{{\mathbf G}}

\def\S{{\mathbf \Sigma}}

\begin{document}

\title{Shapes Characterization on Address Event Representation Using Histograms of Oriented Events and an Extended LBP Approach}
\author[a,b]{Pablo Negri\corref{cor1}\fnref{fn1}}
\ead{pnegri@uade.edu.ar}

\cortext[cor1]{Corresponding author}
\fntext[fn1]{Tel: ++54 11 4000 7307}
\address[a]{CONICET, Godoy Cruz 2290, Buenos Aires, Argentina}
\address[b]{Instituto de Tecnolog\'{i}a, Universidad Argentina de la Empresa (UADE), Lima 717, 1073 Buenos Aires, Argentina}

\maketitle
\begin{abstract}
Address Event Representation is a thriving technology that could change digital image processing paradigm.
This paper proposes a methodology to characterize the shape of objects using the streaming of asynchronous events.
A new descriptor that enhances spikes connectivity is associated with two oriented histogram based representations. 
This paper uses these features to develop both a non-supervised and a supervised multi-classification framework to recognize poker symbols from the Poker-DVS public dataset.
The aforementioned framework, which uses a very limited number of events and a simple class modeling, yields results that challenge more sophisticated methodologies proposed by the state of the art.
A feature family based on context shapes is applied to the more challenging 2015 Poker-DVS dataset with a supervised classifier obtaining an accuracy of 98.5 \%.
The system is also applied to the MNIST-DVS dataset yielding an accuracy of 94.6 \% and 96.3 \% on digit recognition, for scales 4 and 8 respectively.
\end{abstract}


\section{Introduction}

A new paradigm on visual sensing was introduced in 2006 with the first Event-Driven Dynamic Vision Sensor (DVS) \cite{Lichtsteiner:2006}, inspired by the asynchronous Address Event Representation (AER) introduced by Mahowald \cite{Mahowald:1994}, and Kramer's transient detector concept \cite{Kramer:2002}.
The Dynamic Vision Sensor (DVS), also known as \textit{silicon-retinae}, consists of a 128x128 pixel grid, that captures asynchronous individual light changes at the focal plane, reproducing the behavior of biological retinas \cite{Lichtsteiner:2008}.
Each pixel on the sensor operates independently and asynchronously, capturing light changes in low latency and high dynamic.
When such an event occurs, it is transmitted as an information tuple, indicating pixel position on the grid, time stamp and polarity of the event.
Thus this sensor transmits a continuous flow of new events instead of a 2D frame, making it possible to detect event spikes of 10 microseconds (or less), which makes it 100 times faster than high speed conventional cameras (120 fps).
This kind of vision sensors is considered as ``frameless'' providing asynchronous high temporal resolution data.
The present paper proposes a methodology to characterize shapes using the flow of DVS asynchronous events based on spatio-temporal features.

\begin{figure}[h!]
\centerline{
\subfigure[]{\includegraphics[width=50mm]{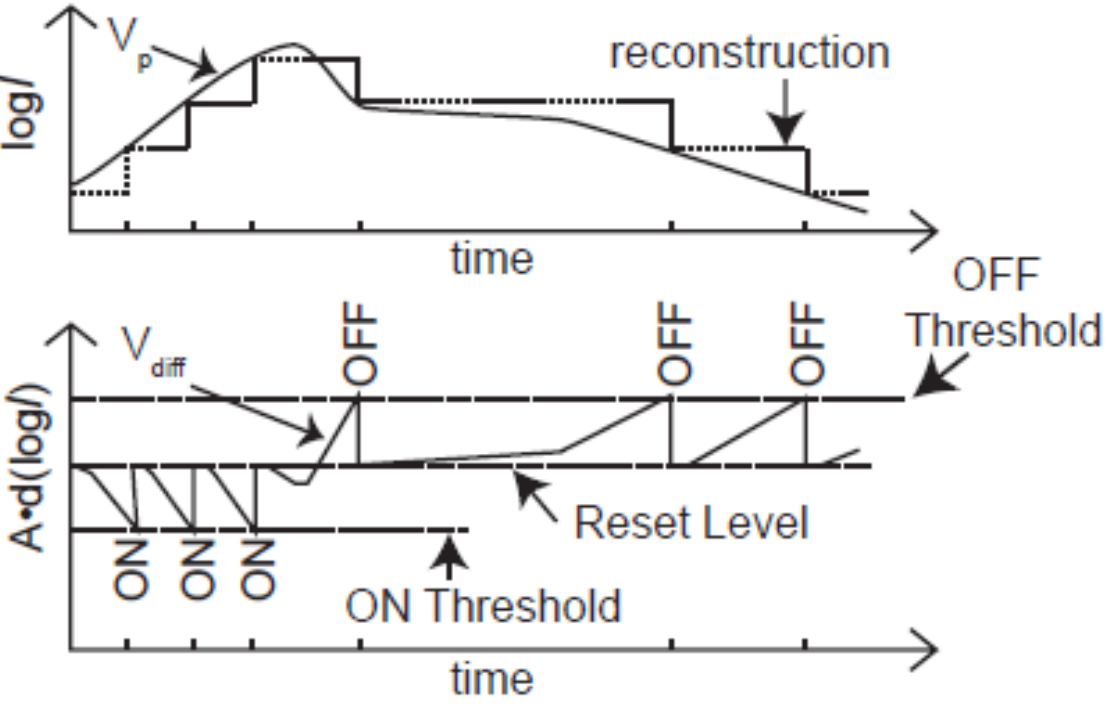}
\label{fig:dvsevent}}
\subfigure[$E_{100}$]{\includegraphics[width=17mm]{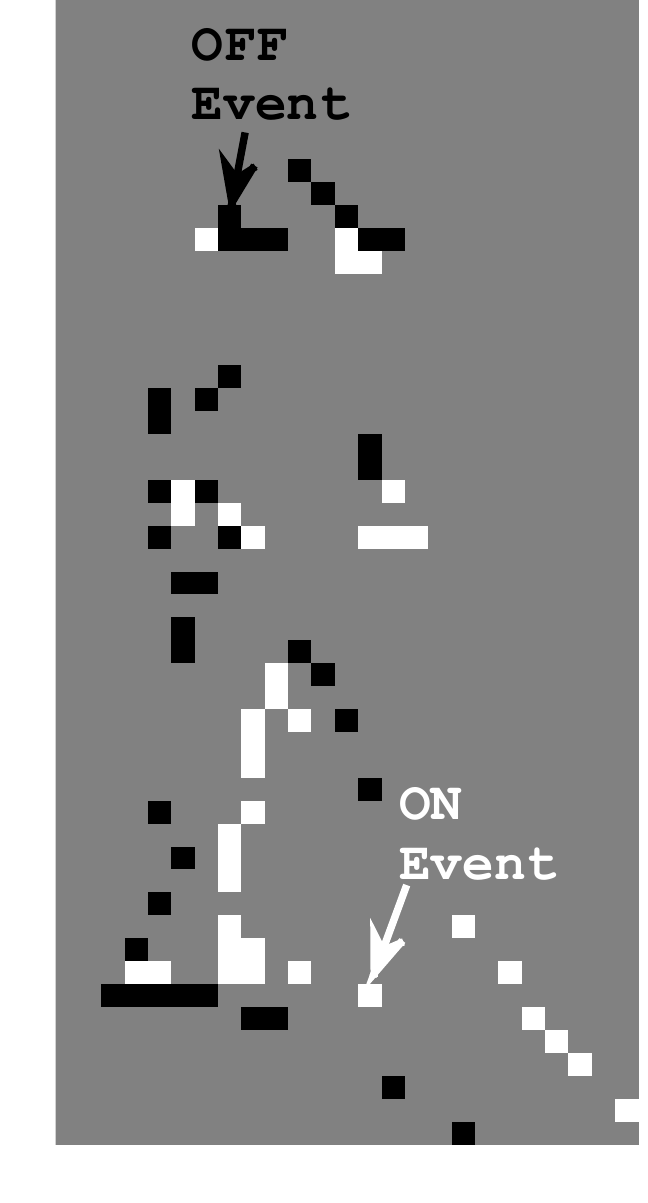}
\label{fig:dvsped100}}
\subfigure[$E_{300}$]{\includegraphics[width=17mm]{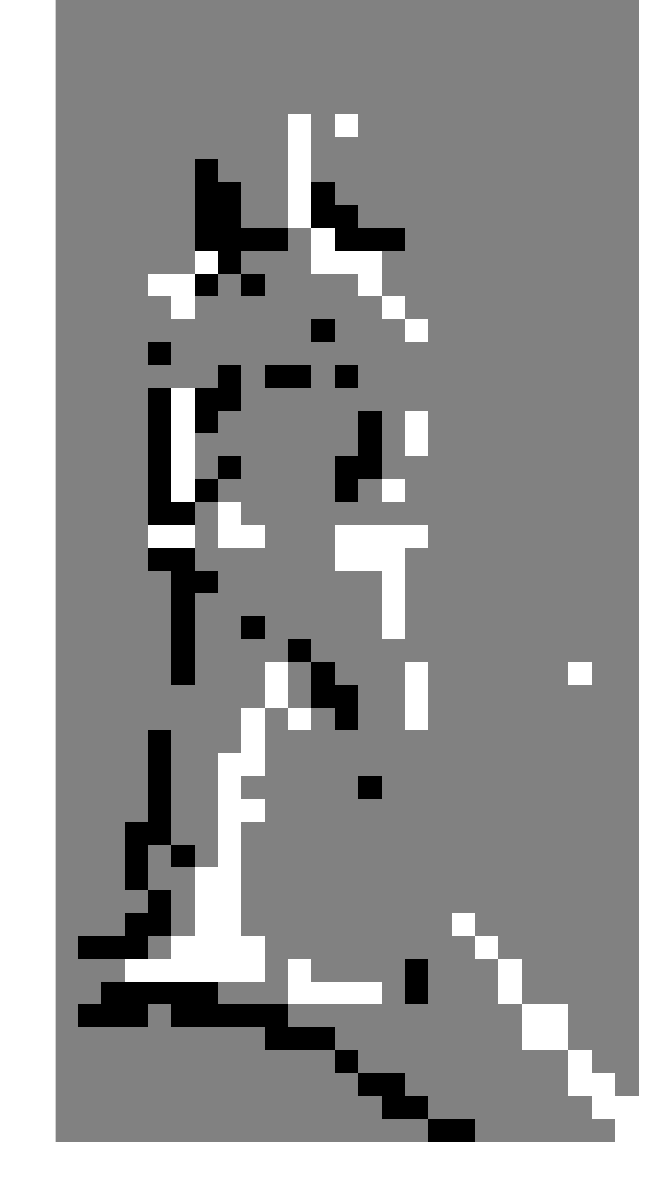}
\label{fig:dvsped300}}
\subfigure[$E_{500}$]{\includegraphics[width=17mm]{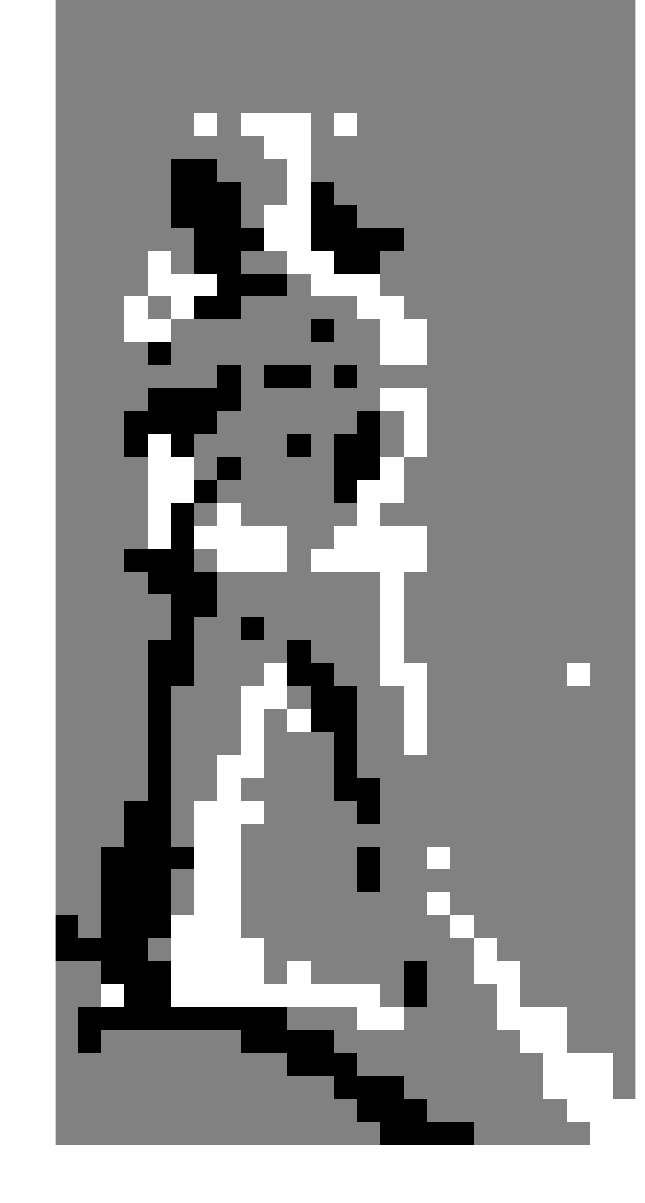}
\label{fig:dvsped500}}
}
\caption{(a) DVS principle of ON and OFF polarity event generation from \cite{Lichtsteiner:2008}, (b) ``Street scene with cars and people walking'' surveillance sample dataset downloaded from \cite{jaer:dataset}.}
\label{fig:dvsoperation}
\end{figure}

Fig. \ref{fig:dvsevent} shows the operation principle diagram of the DVS Address-Event Representation from \cite{Lichtsteiner:2008}.

Each illumination change on the 128x128 pixel grid generates a spiking event $\e=((x,y),t,pol)$, $(x,y)$ being the coordinates of the pixel on the grid, $t$ the event time stamp, and $pol$ the polarity.
Polarity is a binary ON/OFF output. 
ON polarity informs an illumination increase, and OFF polarity is obtained when illumination decreases, as represented in Fig. \ref{fig:dvsevent}.
An event flow composed of $N$ consecutive events is defined as: $\w_N = \{\e_1, ... , \e_N \}$.
The events of $\w_N$ can be mapped on a 2D matrix $E_N$, as shown in Figs. \ref{fig:dvsped100}, \ref{fig:dvsped300} and \ref{fig:dvsped500}, which correspond to the ``Street scene with cars and people walking'' surveillance sample dataset downloaded from \cite{jaer:dataset}.
In \cite{Perez:2013}, this operation is denominated ``histogramming'' the events on $\w$.
$E_{N}(x,y)=1$ if pixel $(x,y)$ corresponds to an incoming event with an ON polarity and it is colored white.
$E_{N}(x,y)=-1$ if polarity of the event at pixel $(x,y)$ is OFF and it corresponds to black dots.
The rest of the pixels of $E_{N}$ are set to zero.

In Fig. \ref{fig:dvsped100}, the number $N$ of events lying on the pedestrian region is set to 100.
The time elapsed between $\e_1$ and $\e_{100}$ defines a temporal window of 13.24 milliseconds (ms).
As can be observed, available information is not enough to recognize the person.
Fig. \ref{fig:dvsped300} and Fig. \ref{fig:dvsped500} show event windows $\w$ with $N=300$ and $N=500$  events.
They are associated with temporal windows of 40.7 ms and 67.7 ms, respectively. 
The number of events now captures sufficient data to identify the moving object as a person.

This simple example raises interesting questions regarding, for instance, the number of events that may be sufficient to recognize an object and whether traditional methodologies can be adapted to take advantage of the new nature of the information.

\subsection{Related Work}

The DVS sensor has received attention from the computer vision community and different kinds of procedures have been proposed to detect objects and recognize shapes using oriented filters, spiking neural networks configurations, or simple histograms.

In \cite{bauer2007embedded}, the low latency of the DVS was utilized to capture highway traffic flow.
Moving vehicles were detected by accumulating the spikes lying on a region of interest within a window of 10 ms and stored in a buffer.
The presence of a vehicle was validated by an established threshold in the number of activated pixels.

Chandrapala \& Shi \cite{chandrapala2016invariant} tackled object recognition based on a feature extractor on a multilayer architecture.
When local activity was detected in a sub-region of the 2D camera grid, the signal was encoded by a Markov chain.
This representation produces a result similar to that of Gabor filters which was sent to the next pyramidal level.
It grouped the data and encoded it in complex features, such as corners or edges.
Neural networks were then trained to recognize different kinds of objects.
Valeiras \textit{et al. }\cite{reverter2015asynchronous} introduced Event-based Gaussian Trackers that approximated the event clouds caused by a moving object using a bivariate Gaussian distribution, making it possible to track simple grid configurations of points for face pose estimation.

Event-driven convolution modules \cite{Serrano-Gotarredona:2008} were employed to implement banks of Gabor filters which captured the orientation of object edges.
They represented the input of Spiking Neural Networks (SNN) architectures for object recognition \cite{Perez:2013,Orchard:2015,Zhao:2015} and were also used in stereo calibration and reconstruction \cite{Camunas:2014}.

Lagorce \textit{et al.} \cite{Lagorce:2016} generated time-surface prototype features from spatio-temporal event-driven clouds.
They employed a hierarchical model architecture (HOTS) consisting of several consecutive layers of increasing detail and including a histogram representation of time-surface activations for each object class.

Recently, Clady \textit{et al.} \cite{Clady:2017} proposed a hand-gesture recognition framework using a histogram representation of flow motion vectors. 
Histograms had a polar grid shape and accumulated the velocity vectors of each event as a global representation of the scene dynamic. 
Their classification algorithm applied Adaboost to select representative histograms, and Bayesian filtering to recognize simple hand movements.

It is also interesting to analyze how frame-based systems compute spatio-temporal features in video sequences.
In general, they generate a 3D representation by joining consecutive frames, which can be considered a temporal sampling of the visual information and depends on the $fps$ rate of the camera.
Using this architecture, different features have been proposed: intensity differences \cite{Viola:2005} and derivatives of optical flow \cite{Dalal:2006} for object recognition; the combination of boundary motion and optical flow histograms for action recognition \cite{Wang:2013}; or scene recognition using 2D and 3D Gaussian filters \cite{Feichtenhofer:2016}.
In \cite{Wan:2014,Wan:2016} they incorporated video depth n the construction of spatio-temporal scale invariant representation features for gesture recognition in RGB-D sequences.
\subsection{Proposed Feature Family}
 
Descriptors transform visual information by evaluating relationships between neighboring pixels.
These associations are organized in mathematical representations as filter outputs, histograms, etc.
Histograms are a powerful tool to characterize objects shape \cite{Dalal:2005}, identify image keypoints \cite{lowe2004distinctive}, or model texture \cite{Ojala:2002}.
They are mainly based on the use of intensities and their gradient calculation.

Because DVS does not provide pixel intensities or colors (new versions of DVS devices will supply this information), special methodologies must be implemented to study pixel/event relationships.
Considering that an isolated event by itself, does not provide useful information, a descriptor using DVS information could analyze the presence of activated neighbors around an activated pixel.
In this way, the event flow is evaluated in relation to neighboring events that are temporally coincident, and belong to the same object.
This paper proposes a methodology to build and to combine special relationships for the data flow using event orientation and event connectivity on histogram representations.

The first procedure is inspired by the Histogram of Oriented Level Lines (HO2L) computed on the Movement Feature Space (MFS) \cite{Negri:2012,Negri:2014}.
Similar to the MFS, incoming flow events only correspond to the boundaries of moving objects, while static objects do not generate them.
Neighboring events that coincide temporally are accumulated in histogram bins, where each bin corresponds to one event orientation.
This local spatio-temporal descriptor is referred to as Histogram of Oriented Events (HOE).
A second approach builds histograms called Events Shape Context (ESC) by evaluating event orientation and distance measurements inspired by Belongie's work \cite{belongie2002shape}, but using a 3D representation.

The event neighborhood analysis employs a modified version of the Local Binary Pattern (LBP) \cite{Ojala:2002} operator called Extended Local Binary Patterns (eLBP).
On gray scale images, LBP performs a simple analysis (binary) about the relationship between gray scale values of neighboring pixels.
In detection problems, these features perform well due to their tolerance to monotone illumination changes \cite{Martinez:2012}.
The eLBP operator,first introduced in \cite{Negri:2016}, seeks to characterize the neighborhood of events in order to enhance edge configurations and penalize isolated events on HOE and ESC representations.

\begin{figure}[h!]
\centerline{
\includegraphics[clip,width=0.65\columnwidth]{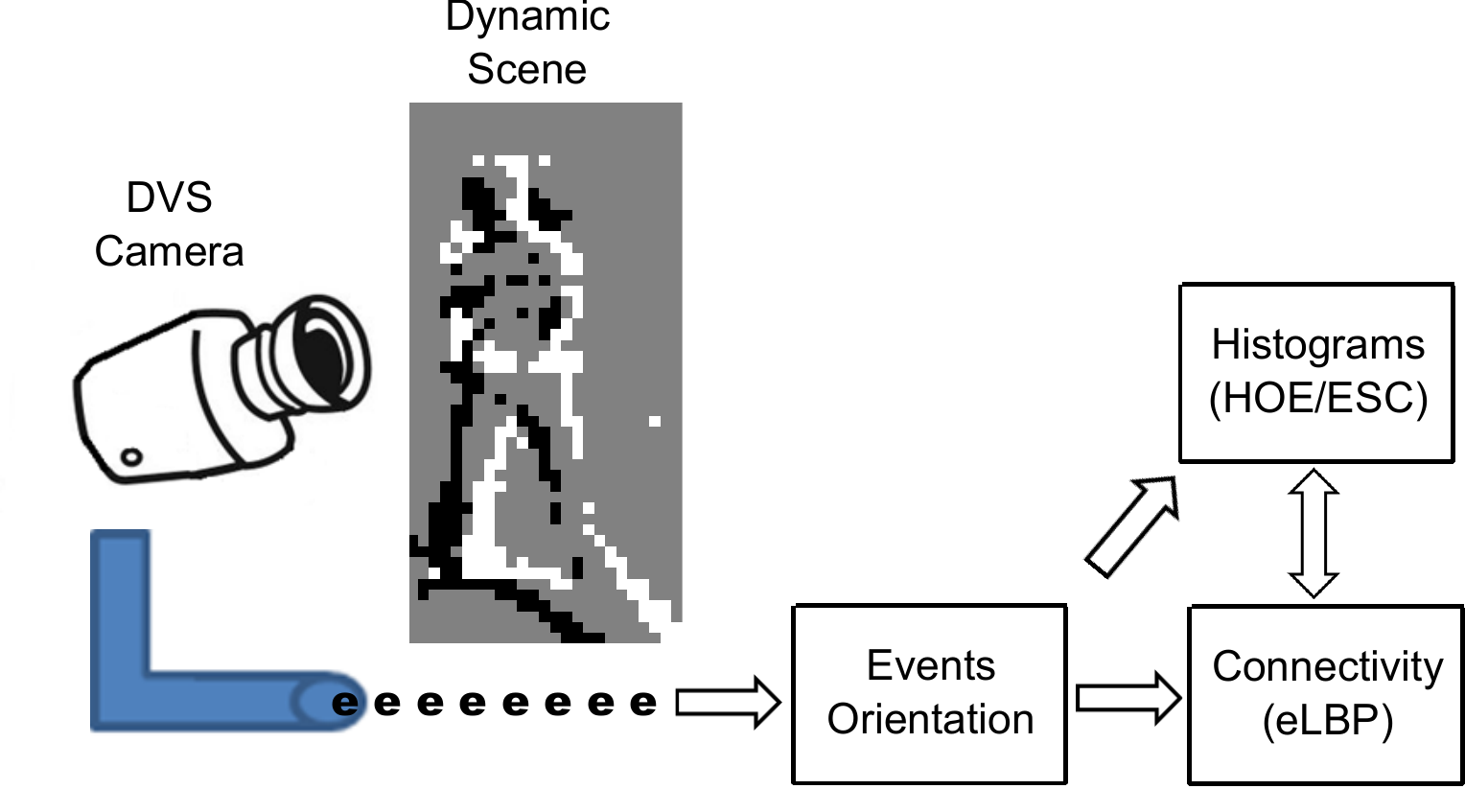}}
\caption{The image shows the HOE, ESC and eLBP feature extraction pipeline.}
\label{fig:featPipeline}
\end{figure}

Fig. \ref{fig:featPipeline} presents the pipeline of the feature extraction algorithm from the flow event data.
In this paper, the potential of the three descriptors, HOE, ESC and eLBP, is evaluated in terms of characterization and recognition of real object shapes captured by DVS.
To this end, two multi-class classification frameworks are implemented to recognize the four classes of the POKER-DVS  \cite{Serrano:2015}: a non-supervised methodology and a supervised classification.
The latter is also applied to the more challenging MNIST-DVS \cite{Serrano:2015}, which is an event-based representation of the original MNIST dataset \cite{LeCun:1998}.

Next section details histogram feature families and shape characterization.
Section \ref{sec:classification} develops the classification frameworks and results are analyzed in section \ref{sec:results}.
Finally, conclusions are discussed in section \ref{sec:conclusions}.
\section{Feature Extraction}

\subsection{Event flow Analysis}

\begin{figure}[h!]
\centering{
\subfigure[]{\includegraphics[clip,width=0.40\columnwidth]{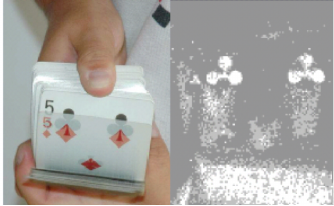}}
\subfigure[One club sign event flow]{\includegraphics[clip,width=0.45\columnwidth]{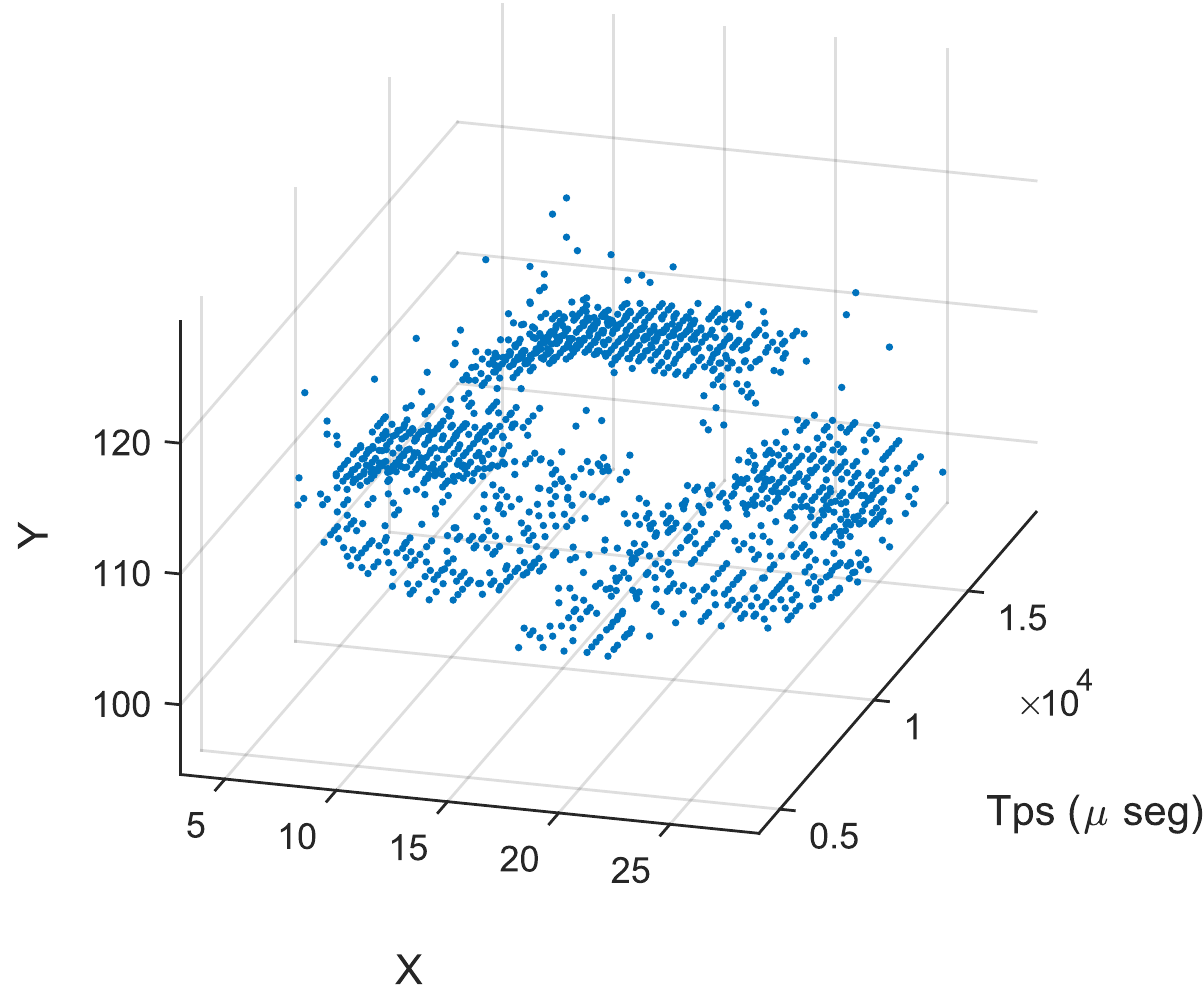}}
}
\caption{Poker-DVS dataset extraction. Left image shows a color framed RGB capture (image from \cite{Perez:2013}), and right image shows the DVS retina captured events of a \textit{club} sign.}
\label{fig:browsing}
\end{figure}

The original Poker-DVS dataset, which was kindly provided by Dr. Bernab{\'e} Linares-Barranco \cite{Perez:2013,Serrano:2015}, is employed to illustrate the histograms extraction methodology on the DVS data.
This dataset shows the use of the DVS retina for very high speed incoming events.
Events flow was obtained by browsing a poker card deck in front of the camera, as shown in Fig. \ref{fig:browsing}.
Several browsing recordings were performed and the dataset consists of the 40 best-looking symbols captured among the records.
Thus, the dataset stored all the events belonging to one symbol while it was visible, until the card disappeared from sight.

\begin{figure}[h!]
\begin{center}
\begin{minipage}[l]{0.28\linewidth}
\subfigure[Rotating Disc]{\includegraphics[clip,width=0.9\columnwidth]{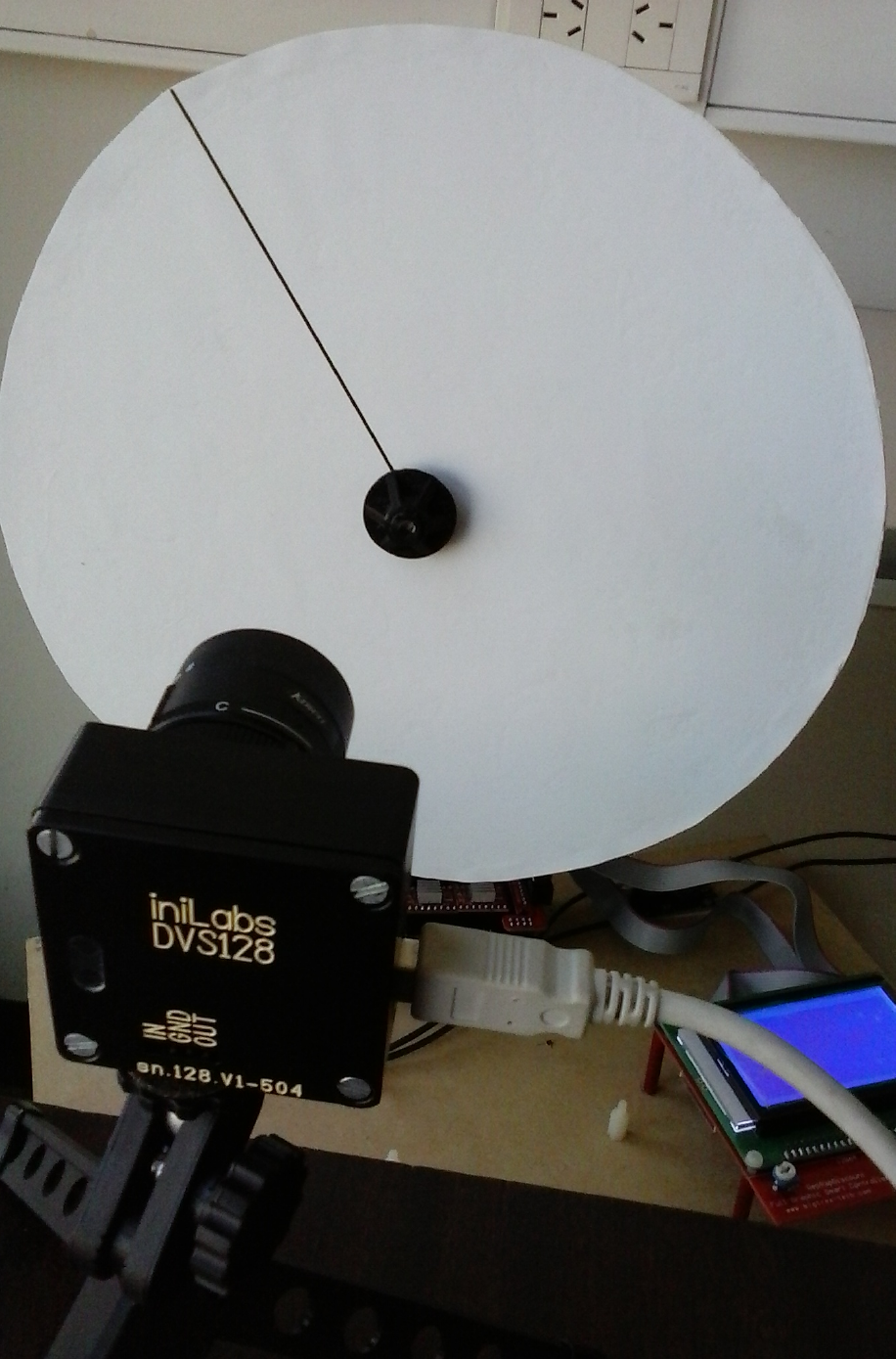}}
\end{minipage}
\begin{minipage}[l]{0.55\linewidth}
\begin{tabular}{cc}
\subfigure[$\omega_1$, $E_{dt=2.5ms}$]{\includegraphics[clip,width=0.40\columnwidth]{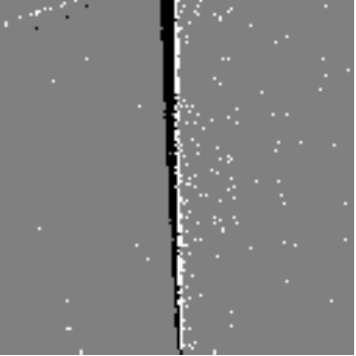}\label{fig:rpm1000dt}} & 
\subfigure[$\omega_3$, $E_{dt=2.5ms}$]{\includegraphics[clip,width=0.40\columnwidth]{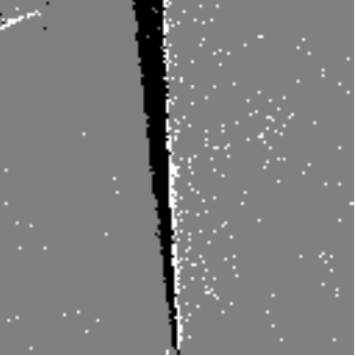}\label{fig:rpm3000dt}} \\
\subfigure[$\omega_1$, $E_{400}$]{\includegraphics[clip,width=0.40\columnwidth]{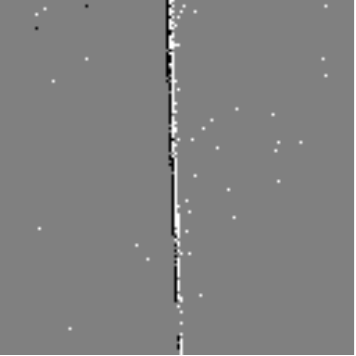}\label{fig:rpm1000N}} & 
\subfigure[$\omega_3$, $E_{400}$]{\includegraphics[clip,width=0.40\columnwidth]{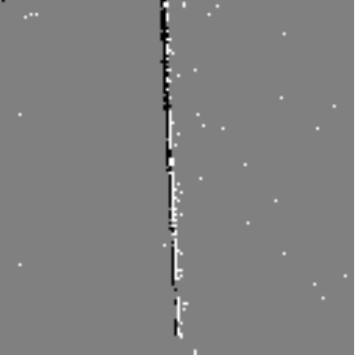}\label{fig:rpm3000N}} \\
\end{tabular}
\end{minipage}
\end{center}
\caption{Rotating bar experiment.}
\label{fig:disco_experiment}
\end{figure}

\begin{lem}\label{lema:1}
The same object passing in front of the retinal camera at different speeds generates similar number of events.
\end{lem}
Lemma \ref{lema:1} states that, theoretically, a moving object with the same trajectory passing in front of DVS will trigger events on the same pixels of the camera.
Therefore, the stream of events will be similar and only the relative time stamps will be different.
Experimentally, the number is different given the stochastic nature of the event generation on the DVS.
To evaluate this error, a rotating bar experiment was implemented as in Benosman's work \cite{Benosman:2013}.
Fig.~\ref{fig:disco_experiment} presents the experiment, where a white disc rotates at three known angular speeds while the bar is captured by the DVS.
The speeds are; $\omega_1=2\pi \ rad \cdot s^{-1}$, $\omega_2= 4 \pi \ rad \cdot s^{-1}$, and $\omega_3=6\pi \ rad \cdot s^{-1}$. 
The average number of events triggered by the rotating bar decreases from 341 for $\omega_1$ to 257 for $\omega_4$.
While the difference between speeds $\omega_1$ and $\omega_2$ is about 24 \%, it is 7 \% for $\omega_2$ and $\omega_3$, and the number of triggered events remains stable for higher speeds.
However, the saturation of the USB cache produces a loss of events when they corresponds to objects moving at very high speeds.

Fig.~\ref{fig:rpm1000dt} and Fig.~\ref{fig:rpm3000dt} show examples of working with fixed temporal windows of $2.5ms$. 
As can be seen, the bar rotating at $\omega_3$ accumulates 1000 events more than the bar rotating at $\omega_1$, deforming the shape.
Because histogram representations are local descriptors, this kind of behavior changes the appearance of the object.
The approach using temporal windows is not suitable to be used with the proposed features families.

Based on Lemma \ref{lema:1}, event flow data is analyzed using \textit{blocking into frames}, or \textit{framing}, which is a well-known methodology employed in speech recognition \cite{rabiner1989tutorial}.
Applied to DVS data, \textit{framing} is modeled by the windows $\w_N=\{\e_1, ... , \e_N \}$ and is employed to characterize an object shape.
\textit{A priori} knowledge about the non-deformable object to be detected allows to define a proper value for $N$. 
In the example of Fig.~\ref{fig:disco_experiment}, the bar has $128$ pixels of height, so both sides should have at least $2 \times 128$ events.
Increasing this quantity by 30 \% to be robust against noise, final $N$ can consist of about 400 events.
Fig.~\ref{fig:rpm1000N} and Fig.~\ref{fig:rpm3000N} show an example of two events windows with $N=400$.
In this way, shape does not change when the bar is rotating at two different speeds.
Additionally, temporal information could also be exploited because it is always saved on $\e_i$.

\subsection{Histograms of Oriented Events}\label{sec:hoe}

In order to describe the shape of one symbol, the orientation of the events is first computed following the procedure of the fitting plane algorithm proposed on \cite{Benosman:2013}, within the set of events $\e_n=((x,y),t,pol) \in \w_N$.
Benosman \textit{et al.} \cite{Benosman:2013} proposed the Event-Based Visual Motion Flow, which considered a neighborhood $\Omega_{\e}$ defined by an $L \times L$ pixel size region around an incoming event $\e$ to fit a plane on a spatio-temporal axis using a regularization method.
The direction of the vector normal to the resulting plane defines the orientation $\theta_n$ and the amplitude of the event.
This approach gives a notion of the flow motion on the scene, but it is not employed to describe objects.
Clady \textit{et al.} \cite{Clady:2015} extend Benosman's approach to detect corners consisting of those event positions where at least two valid fitting planes intersect.
They use a maximum number of events within the vicinity of event $\e$, instead of a temporal window.
This interesting approach inspired Lema \ref{lema:1}, because using this methodology the computation is independent of the speed of the object.

Orientations $\theta_n$ with original values between $0$ and $2\pi$ are converted to \textit{directions} ($0-\pi$) and quantified in $V$ integer values.
In this paper, $V=4$, corresponding to vertical, horizontal and the two diagonals directions.
Each event  $\e_n$ receives an integer value $\Delta_n$ between $1$ and $V$.

\begin{figure}[h!]
\centering{
{\includegraphics[clip,width=0.60\columnwidth]{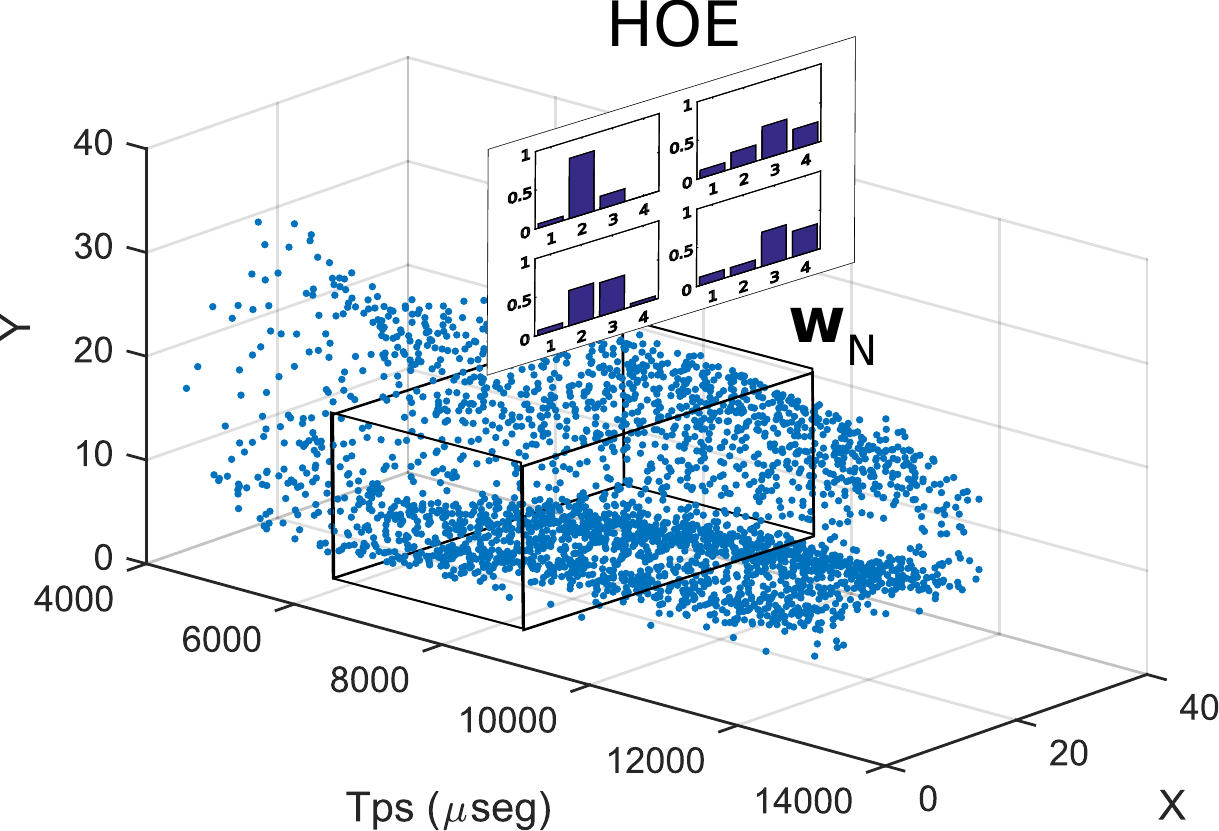}}\\
{\includegraphics[clip,width=0.80\columnwidth]{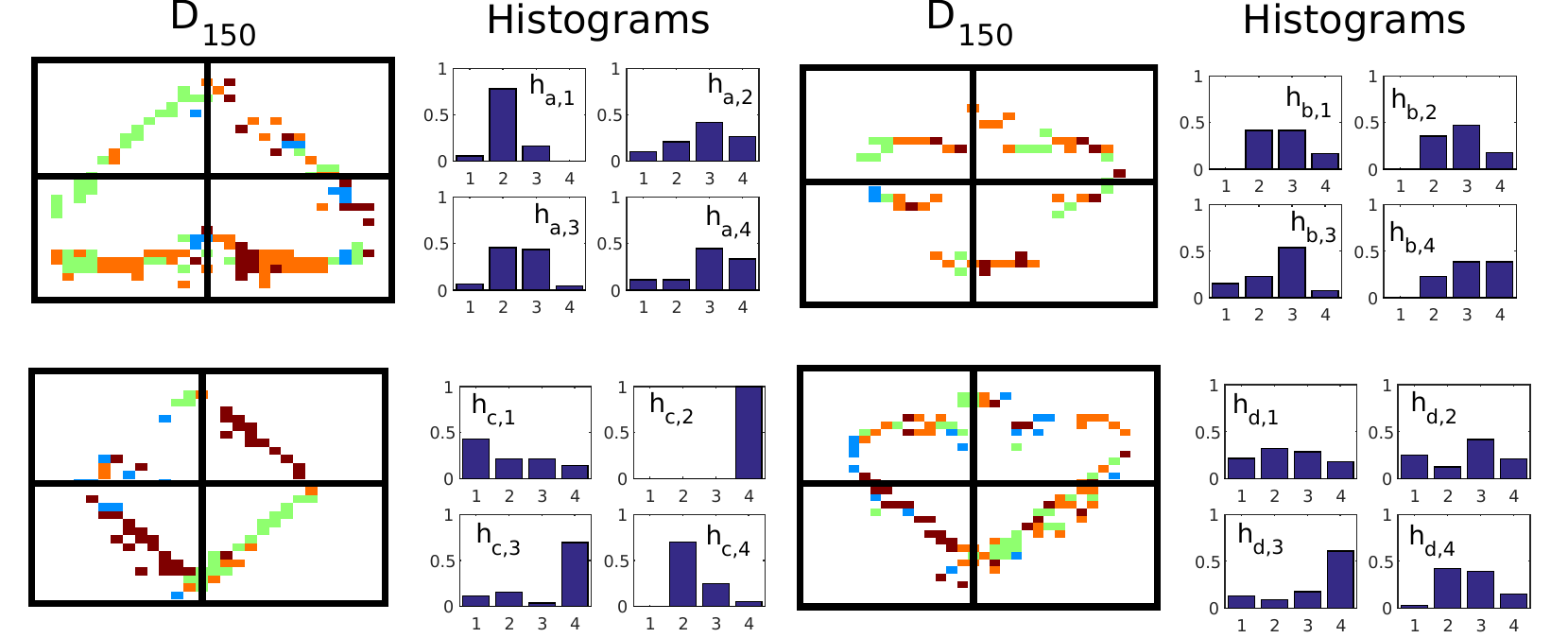}\label{fig:featuresGrids}}}
\caption{First line shows spikes events on the spatio-temporal axis, and a representation of a box, containing events in $\w$, and their corresponding HOE features. Second line is a $D_{150}$ representation for different $\w^a_N$, $\w^b_N$, $\w^c_N$, and $\w^d_N$ event windows and their corresponding histogram representation using $V=4$ directions computed on each cell of the rectangular grid (best seen with colors).}
\label{fig:PokerHO2L}
\end{figure}

Fig. \ref{fig:PokerHO2L} shows four event windows, $\w^a_N$, $\w^b_N$, $\w^c_N$, and $\w^d_N$, one for each poker symbol, with $N=150$.
Each color represents one different event direction.
It is interesting to notice that the edges of the heart and the diamond signs are not uniform and noisy. 
Actually, there are no events generated at those edges because the signs have a motion that is tangential to the edges of the object. 
This happens on both signs and more remarkably on the diamond sign.

To obtain local histogram features, the spatio-temporal region of interest is divided into four quadrants.
Then, Histograms of Oriented Events are computed as detail algorithm \ref{alg:hoe}.
For event window $\w^i_N$, the output of algorithm \ref{alg:hoe} corresponds to $\h_{i,j}$ which is obtained by accumulating the events in the $j$ cell by their direction value.
The values of $\h_{i,j}$ are then normalized to ensure that $\sum_n^V \h_{i,j}(n) = 1$.

\begin{algorithm}[H]
\small
\caption{Histograms of Oriented Events Algorithm}
\label{alg:hoe}
\begin{algorithmic}[1]
\STATE Define a spatio-temporal neighborhood $\Pi$ of set $\w_N$
\STATE Split $\Pi$ into fours quadrants $\kappa_{j=1,...,4}$.
\STATE Initialize fours histograms $\h_{j=1,...,4}$ of $V$ bins with all zero values.
	\FORALL{pairs $(\e_n,\Delta_n)$ of events in $\kappa_j$}
		\STATE $\h_j(\Delta_n) = \h_j(\Delta_n) + 1$
	\ENDFOR
	\FORALL{$\h_j$}
		\STATE $T=\sum_{k=0}^{D-1}\h_j(k)$ 
		\STATE $\h_j = h_j / T$
	\ENDFOR
\RETURN $\h_{j=1,...,4}$
\end{algorithmic}
\end{algorithm}

Finally, a feature vector $\f_i$ concatenating the four histograms $\h_{i,j},j=1,...,4$, that correspond to for each cell of the rectangle containing the temporally coincident events, describes the shape of the object:

\begin{equation}\label{eq:fhoe}
\f^{HOE}=\left[\h_{i,1} \  \h_{i,2} \ \h_{i,3} \ \h_{i,4}\right]
\end{equation}
\subsection{Event Connectivity}

The computation of HOE has a serious drawback on the normalization step.
Isolated events produced by noise can obtain an important value at the histogram deteriorating shape characterization.
This subsection characterizes events spatial neighborhood to enhance histogram representation of connected edges.

\subsubsection{Original LBP}

The $LBP$ operator was initially designed for texture recognition \cite{Ojala:2002}.
It assigns a label to pixel $p$, comparing its intensity value to their 8-connectivity neighbors $q_i$.
The intermediate function $f(p,q_i)$ is defined as:

\begin{equation}\label{eq:lbpfunc}
f(p,q) = 
\left\{ 
\begin{array}{ll}
$1$, & \textrm{if $I(p)-I(q_i)$ $\geq$ $0$} \\
$0$, & \textrm{if $I(p)-I(q_i)$ $<$ $0$} 
\end{array} 
\right.
\end{equation}

\noindent where $I(x)$ refers to the gray scale value at position $x$ of image $I$.
The label returned for the operator is obtained as: $LBP(p) = \sum_{i=0}^8 f(p,q_i) \cdot 2^i$.

Ojala \textit{et al.} \cite{Ojala:2002} incorporated the concept of $transitions$ for each label, which involves the number of changes on the binary string from 0 to 1 and vice versa.
For example, patterns `0000000' (0 transition), `00110000' (2 transit.) and `11000111' (2 transit.) are considered to have a \textit{uniform} appearance of the local binary pattern, describing most frequent features such as edges, corners or spots.
Other binary labels, with more transitions, such as `11011001' (4 transit.) and `01010001' (6 transit.) are considered as not uniform.
Another particularity of the $LBP$ operator is that the patterns are circular, i.e. `00110000' is the same as `11000000'.
In this way, the texture descriptor becomes robust to rotations.

\subsubsection{Extended LBP Characterizing Events Connectivity}

To characterize connectivity between events, equation \ref{eq:lbpfunc} is modified to adapt to DVS data.
Instead of measuring a luminosity change around a pixel, the operator detects if an event is activated within a spatio-temporal space.
The neighborhood around the central point is evaluated using the equality condition:

\begin{equation}\label{eq:elbpfunc}
f(\e_p,\e_q) = 
\left\{ 
\begin{array}{ll}
$1$, & \textrm{if $M(\e_p) = M(\e_{q,i})$} \\
$0$, & \textrm{otherwise} 
\end{array} 
\right.
\end{equation}

$M(\e_p)$ can be modeled as a 2D generic matrix.
If event polarity is considered to analyze connectivity $M(\e_p)$, corresponds to $E_N(\e_p)$.
When connectivity analysis corresponds to the direction of the events, the $D_N(\e_p)$ matrix is employed and the operator is activated if both connected events have the same direction.

\begin{figure}[h!]
\footnotesize
\centering
\begin{tabular}{c@{\hskip 0.03in}c@{\hskip 0.03in}c@{\hskip 0.03in}c@{\hskip 0.03in}c@{\hskip 0.03in}c@{\hskip 0.03in}c@{\hskip 0.03in}c@{\hskip 0.03in}c}
\includegraphics[clip,width=0.09\columnwidth]{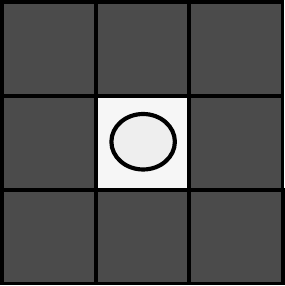} &
\includegraphics[clip,width=0.09\columnwidth]{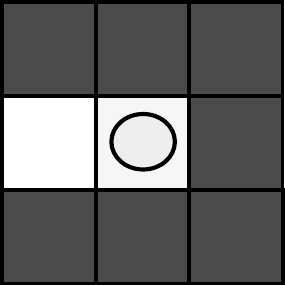} &
\includegraphics[clip,width=0.09\columnwidth]{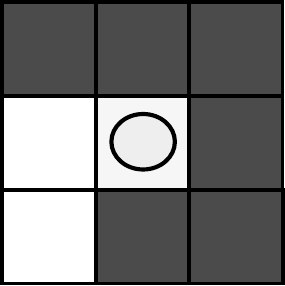} &
\includegraphics[clip,width=0.09\columnwidth]{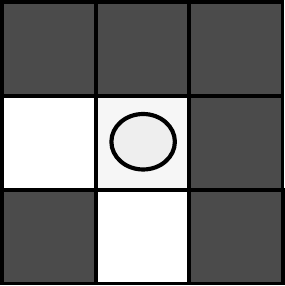} &
\includegraphics[clip,width=0.09\columnwidth]{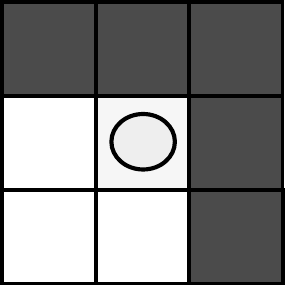} & 
\includegraphics[clip,width=0.09\columnwidth]{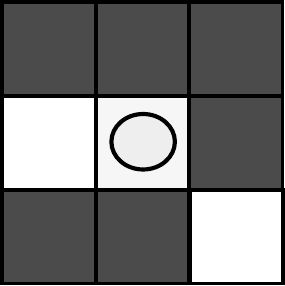} &
\includegraphics[clip,width=0.09\columnwidth]{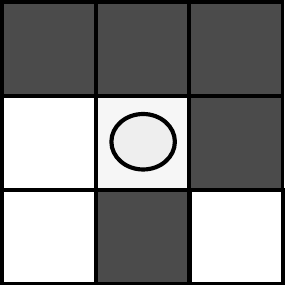} &
\includegraphics[clip,width=0.09\columnwidth]{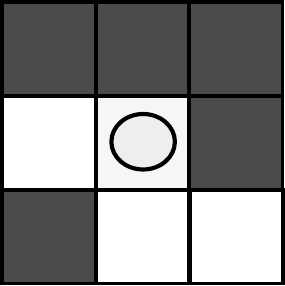} &
\includegraphics[clip,width=0.09\columnwidth]{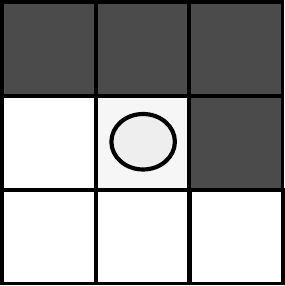} \\
255 - '1' & 254 - '2' & 252 - '3' & 250 - '4' & 248 - '5' & 246 - '6' & 244 - '7' & 242 - '8' & 240 - '9' \\
\includegraphics[clip,width=0.09\columnwidth]{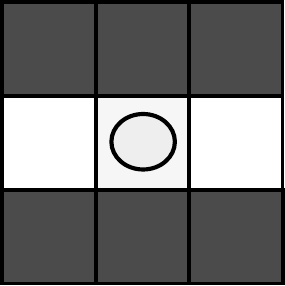} &
\includegraphics[clip,width=0.09\columnwidth]{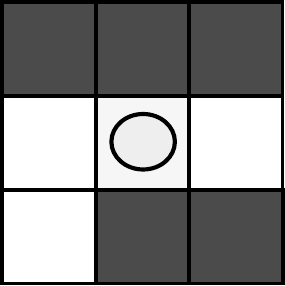} &
\includegraphics[clip,width=0.09\columnwidth]{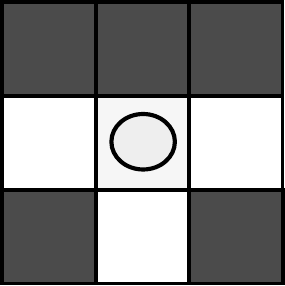} &
\includegraphics[clip,width=0.09\columnwidth]{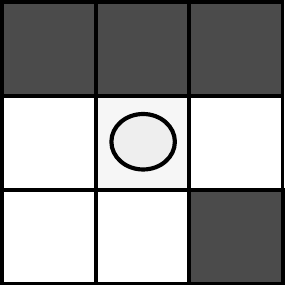} &
\includegraphics[clip,width=0.09\columnwidth]{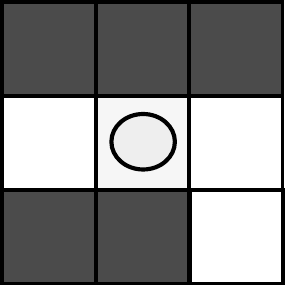} &
\includegraphics[clip,width=0.09\columnwidth]{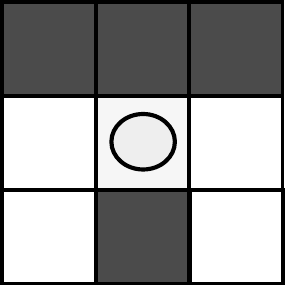} &
\includegraphics[clip,width=0.09\columnwidth]{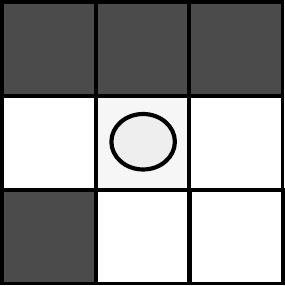} &
\includegraphics[clip,width=0.09\columnwidth]{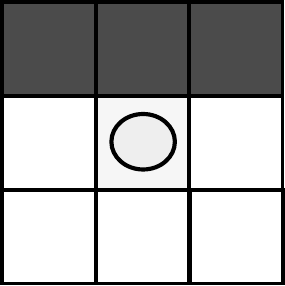} &
\includegraphics[clip,width=0.09\columnwidth]{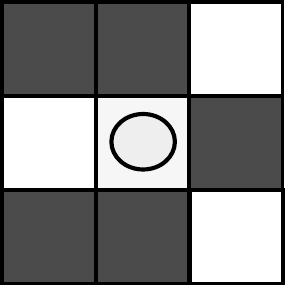} \\
238 - '10' & 236 - '11' & 234 - '12' & 232 - '13' & 230 - '14' & 228 - '15' & 226 - '16' & 224 - '17' & 214 - '18' \\
\includegraphics[clip,width=0.09\columnwidth]{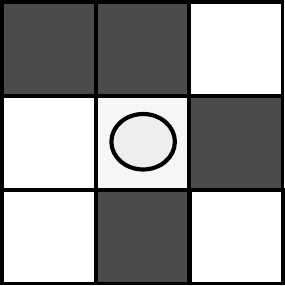} &
\includegraphics[clip,width=0.09\columnwidth]{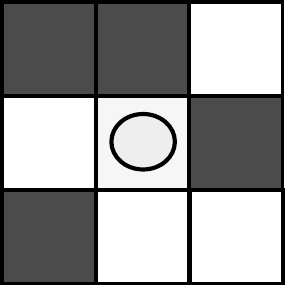} &
\includegraphics[clip,width=0.09\columnwidth]{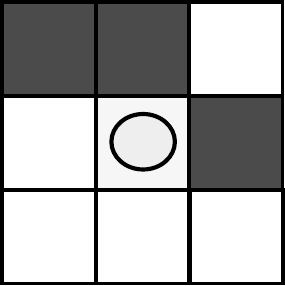} &
\includegraphics[clip,width=0.09\columnwidth]{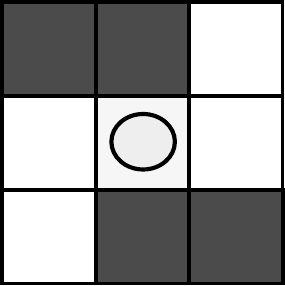} &
\includegraphics[clip,width=0.09\columnwidth]{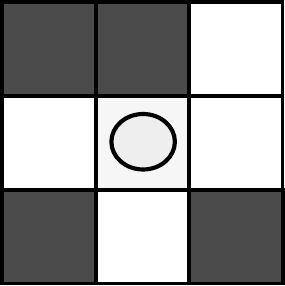} &
\includegraphics[clip,width=0.09\columnwidth]{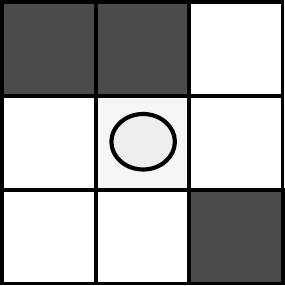} &
\includegraphics[clip,width=0.09\columnwidth]{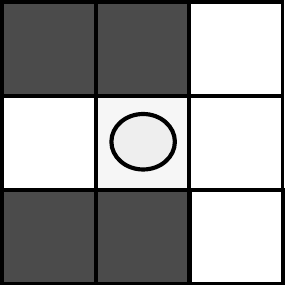} &
\includegraphics[clip,width=0.09\columnwidth]{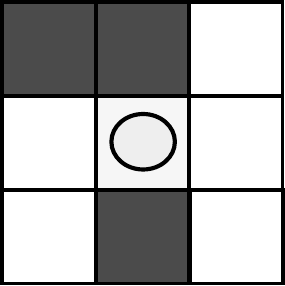} &
\includegraphics[clip,width=0.09\columnwidth]{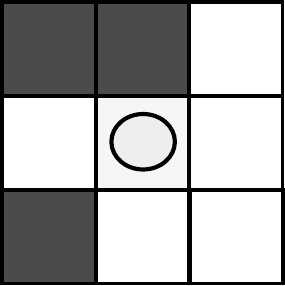} \\
212 - '19' & 210 - '20' & 208 - '21' & 204 - '22' & 202 - '23' & 200 - '24' & 198 - '25' & 196 - '26' & 194 - '27' \\
\includegraphics[clip,width=0.09\columnwidth]{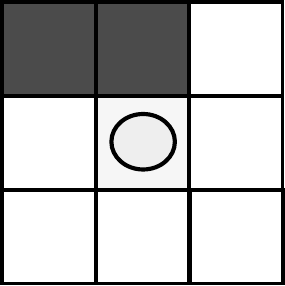} &
\includegraphics[clip,width=0.09\columnwidth]{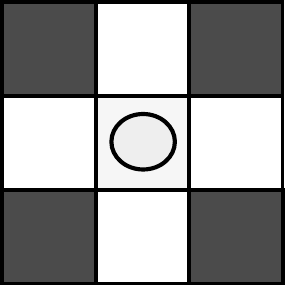} & 
\includegraphics[clip,width=0.09\columnwidth]{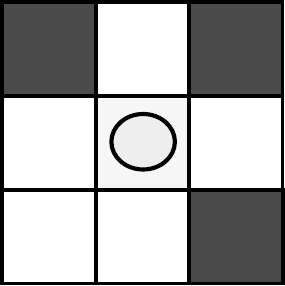} &
\includegraphics[clip,width=0.09\columnwidth]{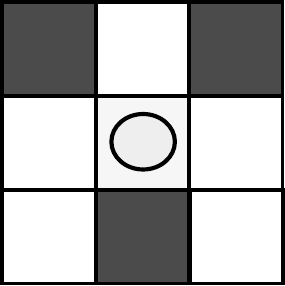} &
\includegraphics[clip,width=0.09\columnwidth]{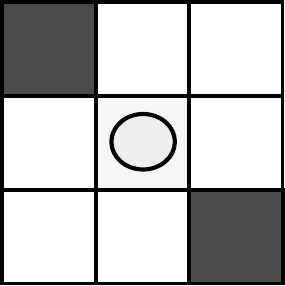} & 
\includegraphics[clip,width=0.09\columnwidth]{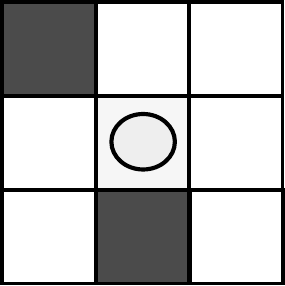} &
\includegraphics[clip,width=0.09\columnwidth]{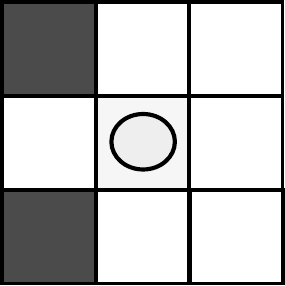} &
\includegraphics[clip,width=0.09\columnwidth]{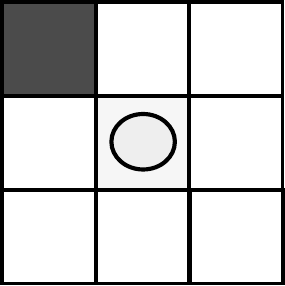} &
\includegraphics[clip,width=0.09\columnwidth]{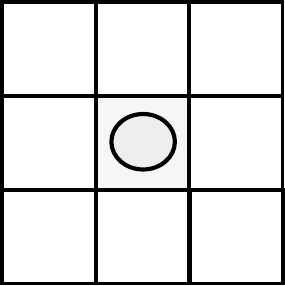} \\
192 - '28' & 170 - '29' & 168 - '30' & 164 - '31' & 136 - '32' & 132 - '33'  & 130 - '34' & 128 - '35' & 0 - '36' \\
\end{tabular}
\caption{Extended LBP patterns to characterize event connectivity. The binary code of each pattern is shown as well as the corresponding eLBP code 'x' to identify the pattern.}
\label{fig:extLBP}
\end{figure}

Fig. \ref{fig:extLBP} presents all the operators $eLBP(\e_p)$ identifying connectivity on DVS events.
Two transition patterns captured information at the end of a segment (patterns $'28'$ and $'35'$), linear edges ($'24'$, $'26'$, $'32'$, $'33'$, $'34'$, etc.) or other edge/region  configurations ($'1'$, $'2'$, $'3'$, $'5'$, $'8'$, $'10'$, etc.).
Four transition patterns characterized different configurations of possible connected edges.
In total, 36 canonical patterns were defined, including isolated events ($'36'$).

Characterization of the event window shape was completed by giving a weight to each event $\e_p$ considering the connectivity code of $eLBP(\e_p)$, as shown in Table \ref{tab:lbpWeights}.
These weights enhanced events with an \textit{edge kind} connectivity, and penalized other event configurations, such as isolated pixels.

\begin{table}[h!]
\scriptsize
\centering
	\begin{tabular}{|l|c|r|}
	\hline
	\textbf{Tag} & \textbf{eLBP codes} & \textbf{Weight} \\
	\textbf{Connectivity} &            & \\
	\hline
	\hline
	LINE & 11,15,16,18,19,21,22,24 & 1.0 \\
	 & 25,26,27,32,33,34 & \\
\hline
	FILL & 1,2,3,5,6,9,12,13, & 0.75 \\
	 & 20,23,29,30,31 & \\
\hline
	LATERAL & 4,7,8,10,13 & 0.75 \\
\hline
	ENDPOINT & 17,35,28 & 0.5 \\
\hline
	ISOLATE & 36 & 0.3 \\
	\hline
	\end{tabular}
\caption{Extended LBP weights.}
\label{tab:lbpWeights}
\end{table}

Codes with the LINE tag got the highest weights, followed by FILL and LATERAL tags. 
Codes with the ENDPOINT tag had a weight of $0.5$, and the ISOLATE tag events obtained the lowest weight.
These weights are employed on histogram $\h_i$, when it is necessary to normalize the bins values.
This time, it was not necessary for the sum of bins on $\h_i$ to equal one ($\sum_k^N \h_i(k) \leq 1$).
Algorithm \ref{alg:hoe_elbp} shows the procedure to compute the weighted histograms using the eLBP approach.

\begin{algorithm}[H]
\small
\caption{HOE with eLBP}
\label{alg:hoe_elbp}
\begin{algorithmic}[1]
\STATE Define a spatio-temporal neighborhood $\Pi$ of set $\w_N$
\STATE Split $\Pi$ into fours quadrants $\kappa_{j=1,...,4}$.
\STATE Initialize fours histograms $\h_{j=1,...,4}$ of $V$ bins with all zero values.
	\FORALL{pairs $(\e_n,\Delta_n)$ of events in $\kappa_j$}
		\STATE Define the eLBP code $\vartheta_n$ of $\e_n$ using eq. \ref{eq:elbpfunc} and a spatio-temporal neighborhood $\Omega_{\e_n}$.
		\STATE Gets weight $\omega(\vartheta_n)$ from table \ref{tab:lbpWeights}
		\STATE $\h_j(\Delta_n) = \h_j(\Delta_n) + \omega(\vartheta_n)$
	\ENDFOR
	\FORALL{$\h_j$}
		\STATE Set $T_j$ as the number of events lying on $\kappa_j$
		\STATE $\h_j = h_j / T_j$
	\ENDFOR
\RETURN $\h_{j=1,...,4}$
\end{algorithmic}
\end{algorithm}

There are two possible analyses for the connectivity of events: the use of polarity or the use of direction.
The former considers the connectivity of neighboring events with the same polarity.
The latter adds the condition that neighboring events have the same direction to be considered connected.

\begin{figure}[t!]
\centering
\includegraphics[clip,width=0.80\columnwidth]{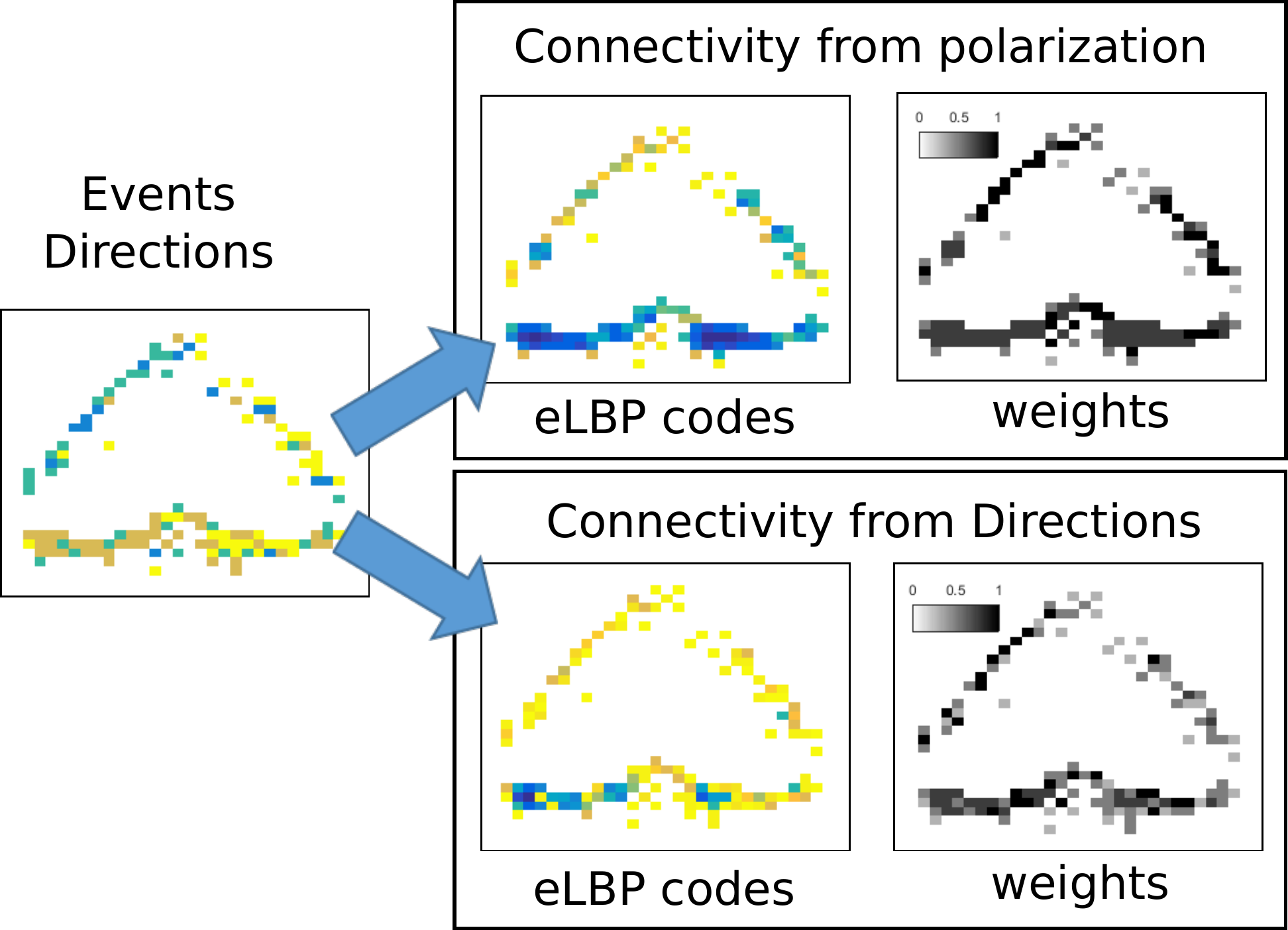}
\caption{Connectivity analysis using the eLBP approach on the events generated by a heart sign (best seen with colors).}
\label{fig:eLBPheart}
\end{figure}

Fig. \ref{fig:eLBPheart} shows an example of an event window corresponding to a heart symbol from the dataset.
On the ``Connectivity from Polarization'' block, the eLBP codes were obtained using the polarity information: $M(\e_p) = E_N(\e_p)$ on equation \ref{eq:elbpfunc}.
In the figure, events were histogrammed on a 2D matrix, and each one had a different color depending on the eLBP code.
These codes were employed to compute a 2D matrix using the weights of table \ref{tab:lbpWeights}.
The example illustrates how isolated events received lower weights while connected events got highest weights and thus enhanced their influence on the HOE computation.
In the following section, the combination of HOE and eLBP analysis using polarity is expressed as HOE+eLBP4Pol. 

On the ``Connectivity from Direction'' block, the direction connectivity was obtained using: $M(\e_p) = D_N(\e_p)$ on equation \ref{eq:elbpfunc}.
The combination with HOE was stated as HOE+eLBP4Dir.
As can be seen, the weights of the events obtained lower values because connected events with same direction were rare in this $\w_N$.

\subsection{Event Shape Context (ESC)}\label{sec:ESC}

\begin{figure}[h!]
\begin{center}
\subfigure[]{\includegraphics[clip,width=0.34\columnwidth]{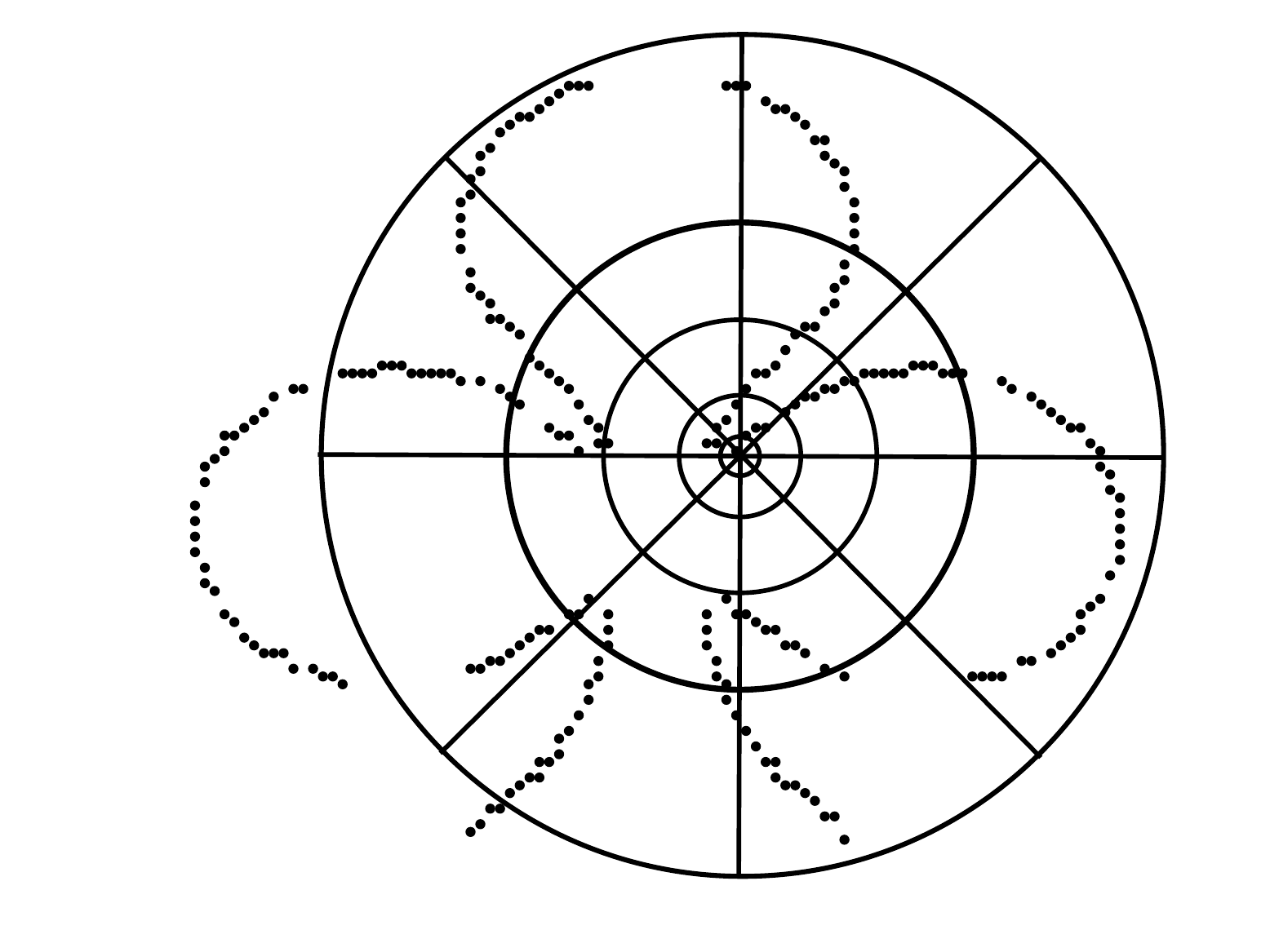}\label{fig:original_sc}}
\subfigure[]{\includegraphics[clip,width=0.44\columnwidth]{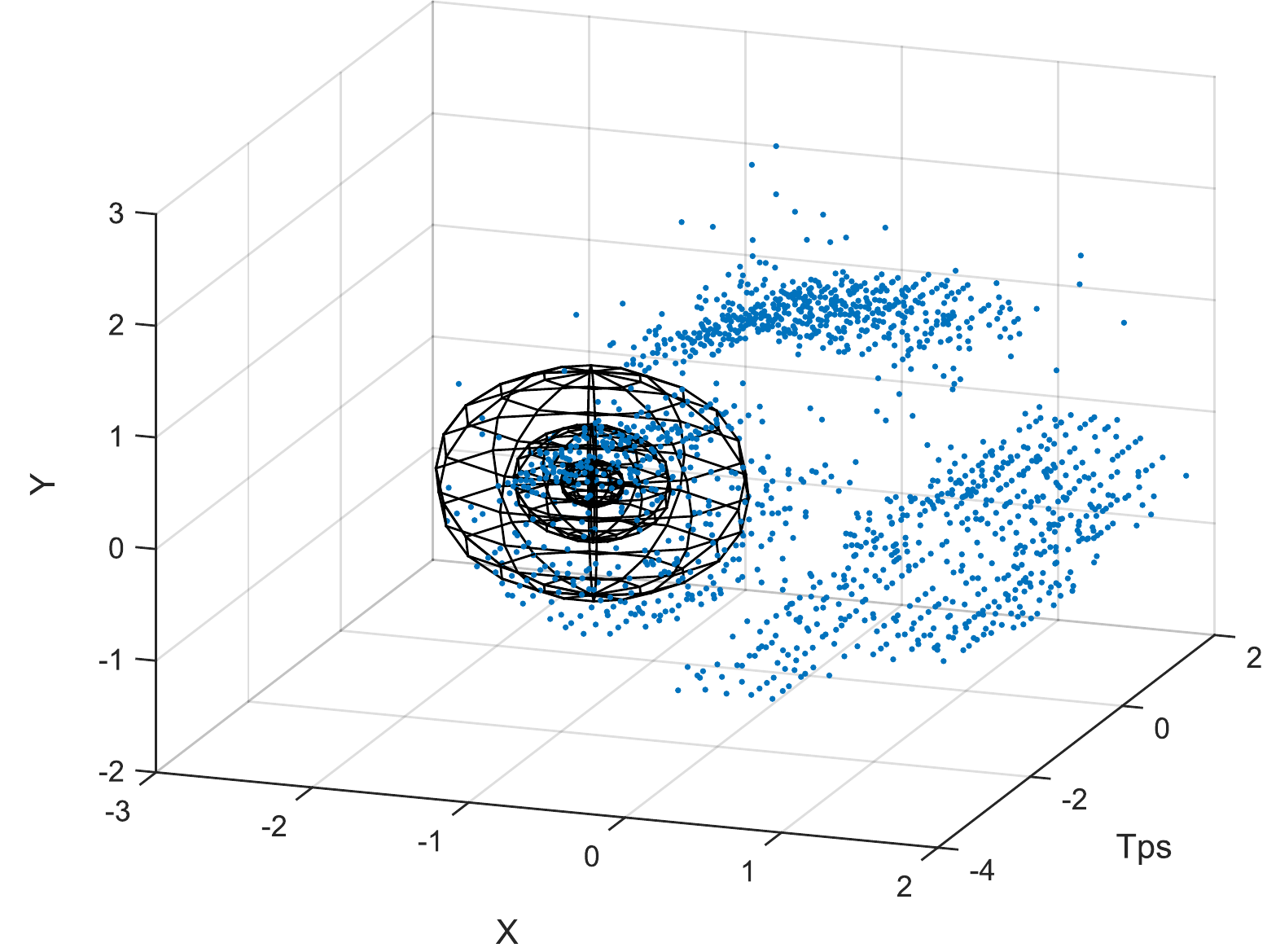}\label{fig:spherical_sc}}
\end{center}
\begin{center}
\subfigure[]{\includegraphics[clip,width=0.24\columnwidth]{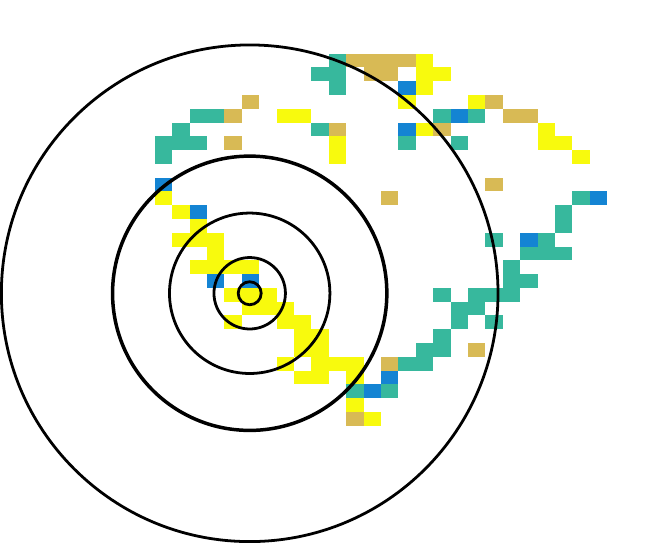}\label{fig:space_sc}}
\subfigure[]{\includegraphics[clip,width=0.24\columnwidth]{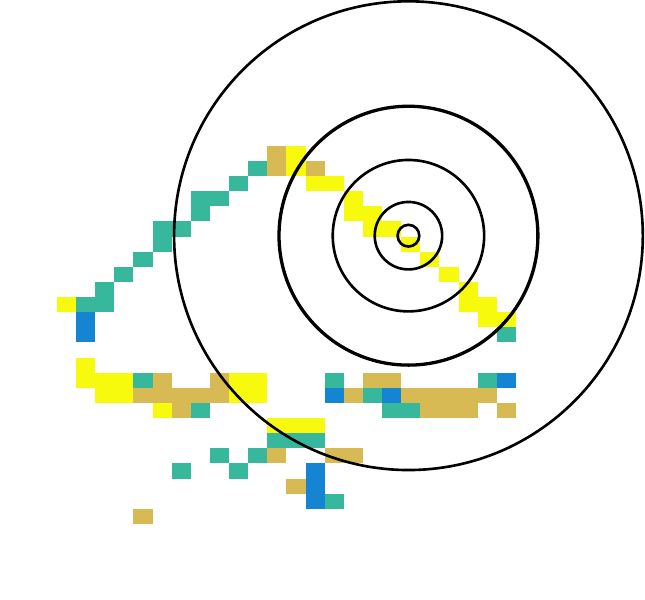}\label{fig:ispade_sc}}
\subfigure[]{\includegraphics[clip,width=0.24\columnwidth]{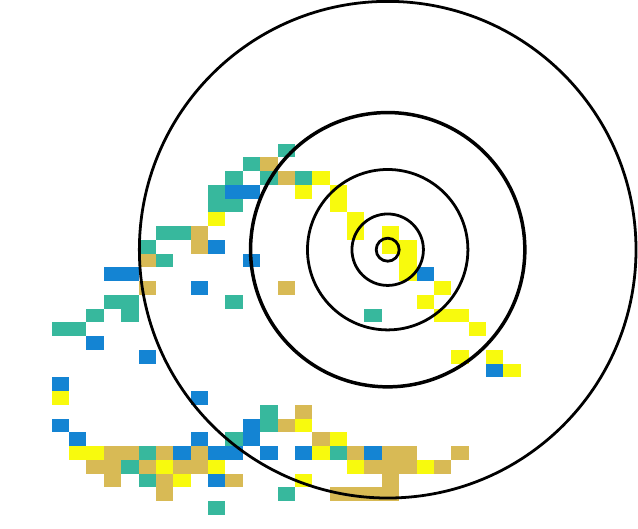}\label{fig:heart_sc}}
\end{center}
\caption{(a) Original shape context grid, (b) proposed spherical Event Shape Context representation, (c) shows a regular \textit{spade}, (c) is an inverted \textit{spade}, and (d) is a \textit{heart} sign. (c), (d) and (e) are presented on 2D for better understanding.}
\label{fig:eSC}
\end{figure}

The 2015 Poker-DVS dataset introduced a new challenge because some cards were inverted, and consequently, the sign was recorded upside-down.
This makes the use of HOE  inappropriate to characterize the shapes because it is a local descriptor.
This section describes another histogram feature family that follows the guidelines of the Shape Context proposed by Belongie \cite{belongie2002shape}.
Fig.~\ref{fig:original_sc} shows the original grid of the Shape Context as a log-polar distribution around an edge point of the object.
It consists of rings divided into cells and centered on the pixel position of interest.
The Shape Context of this point is a histogram where each bin count the number of points lying inside their corresponding cell.
To compare two different shapes, their points' Shape Context are matched in pairs and must satisfy a similarity function.

This paper proposes an Event Shape Context (ESC) descriptor of each event in the window $\w_N$ which is projected to a 3D spatio-temporal representation axis.
To compute the new operator the axis must be normalized to zero mean and unit variance, as shown in Fig.~\ref{fig:spherical_sc}.
The ESC grid is then transformed into a spherical 3D representation.
Additionally, to transform this distribution invariant to the inversion, cells inside the rings are eliminated, as shown in Figs. \ref{fig:space_sc}, \ref{fig:ispade_sc}, and \ref{fig:heart_sc} (grids are presented in 2D for simplicity purposes). 
Finally, the new operator considers pairs $(\e_n,\Delta_n) \in \w_N$, taking into account direction $\Delta_n$ of the events.
In \cite{mori2005efficient} is proven that incorporating direction improves matching results.
Then, each ESC $\s_i$ of event $\e_i$ is composed of histograms that specialize in each direction.

The individual histogram of $\e_i \in \w_N$ specialized in direction $v$ (there are $V$ directions in total) is defined as:

\begin{eqnarray}
s^k_{i,j} & = & \# \{\e_n \neq \e_i : \|\e_n-\e_i\| \in ring(k) \wedge \Delta_n = v \}_{n=1,...,N} \label{eq:histogramorientedshapecontextevent} \\ 
\s_i & = & \left [ s^1_{i,1}, ...,  s^1_{i,V} \ s^2_{i,1}, ...,  s^2_{i,V} \ ... \ s^R_{i,1}, ...,  s^R_{i,V} \right ] \label{eq:orientedshapecontextevent}
\end{eqnarray}

\noindent where bin $k$ is related to the Euclidean distance in the 3D space between $\e_i$ and the other events in $\w_N$ using $\| \|$.
This space is modeled as $R$ rings (spheres) centered around the event and with uniform size in log-polar space, where each bin of $k$ of $s^k_{i,j}$ counts the number of events with direction $v$ lying inside of it.
Thus, there are $R$ histograms $s^k_i$ with $V$ bins $\s^v_i \in \Re^{1 \times RV}$.
Each histogram $s^k_i$ must be normalized in order to assure that $\sum_v s^k_i(v) = 1$.

Equation \ref{eq:orientedshapecontextevent} shows how the individual $s^k_i$ are arranged in the Event Shape Context of $\e_i$.
To describe the shape of all events in $\w_N$ the feature grows to a $NRV$ size.

In order to obtain a more compact characterization of the shape, this paper proposes a methodology to group the $\s_i$ histograms in the following way.
The ESC descriptors of events $\e_i \in \w_N$ having the same direction $\Delta_i=v$ are grouped into an array $\G^v \in \Re^{n_v \times RV}$, $n_v$ being the number of events with direction $v$:

\begin{equation}
\G^v  =  \left( \begin{array}{c} 
\sc^v_1 \\
\sc^v_2 \\
\vdots \\
\sc^v_{n_v}
\end{array}
\right) 
=
\left( \begin{array}{ccccccccccccc} 
s^1_{1,1} & ... &  s^1_{1,V} & \ & ... &\ & s^R_{1,1} & ... & s^R_{1,V} \\
s^1_{2,1} & ... &  s^1_{2,V} & \ & ... &\ & s^R_{2,1} & ... & s^R_{2,V} \\
\vdots & \ddots &  \vdots & \ & \ddots &\ & \vdots & \ddots & \vdots \\
s^1_{n_v,1} & ... &  s^1_{n_v,V} & \ & ... &\ & s^R_{n_v,1} & ... & s^R_{n_v,V} \\
\end{array}
\right) 
\end{equation}

Then, a single histogram $\g^v$ is obtained by computing the average of all the elements in each column of $\G^v$:

\begin{equation}
\g^v  = \frac{1}{n_v}\left [ \begin{array}{ccccccccccccc} 
\sum_{i=1}^{n_v}  s^1_{i,1} & ... & \sum_{i=1}^{n_v}  s^1_{i,V} & ... & 
\sum_{i=1}^{n_v}  s^R_{i,1} & ... & \sum_{i=1}^{n_v}  s^R_{i,V} \\
\end{array} \right ] \label{eq:meanSC}
\end{equation}

\begin{figure}[h!]
\begin{center}
\subfigure[$\f^{ESC}$ of regular \textit{spade} sign]{\includegraphics[clip,width=0.45\columnwidth]{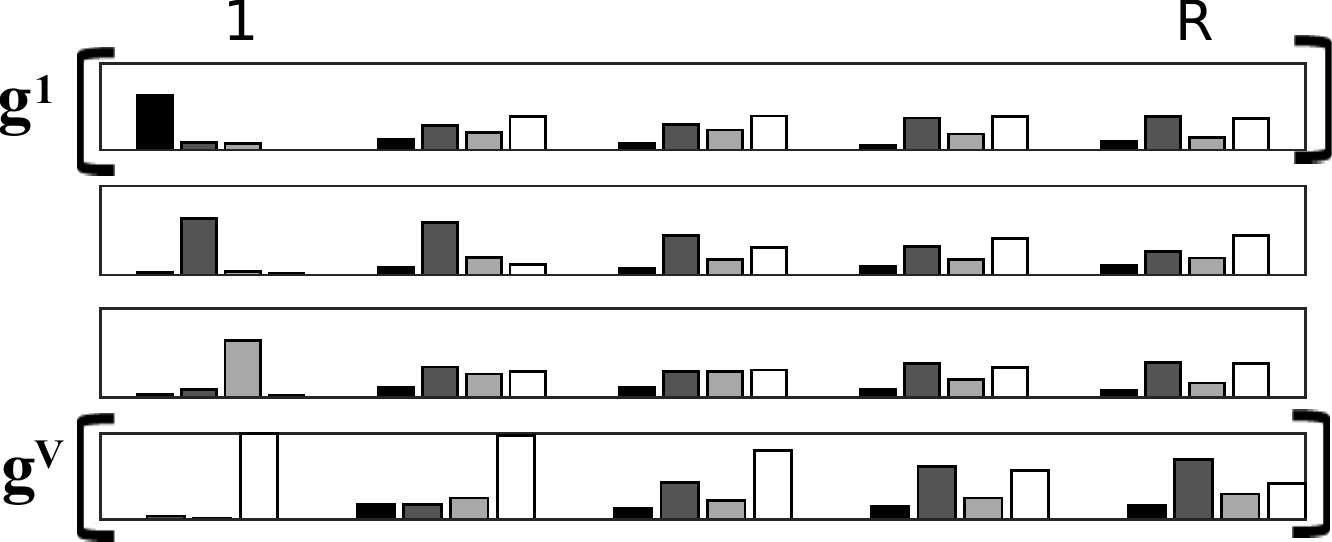}} 
\subfigure[$\f^{ESC}$ of inversed \textit{spade} sign]{\includegraphics[clip,width=0.45\columnwidth]{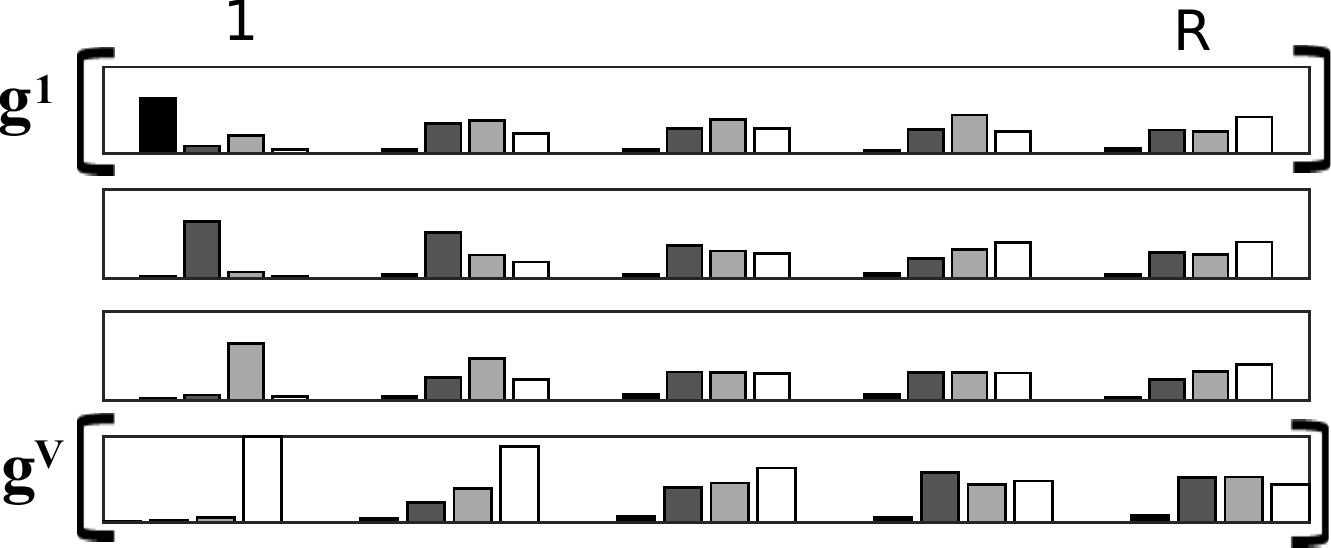}} \\
\subfigure[$\f^{ESC}$ of \textit{heart} sign]{\includegraphics[clip,width=0.45\columnwidth]{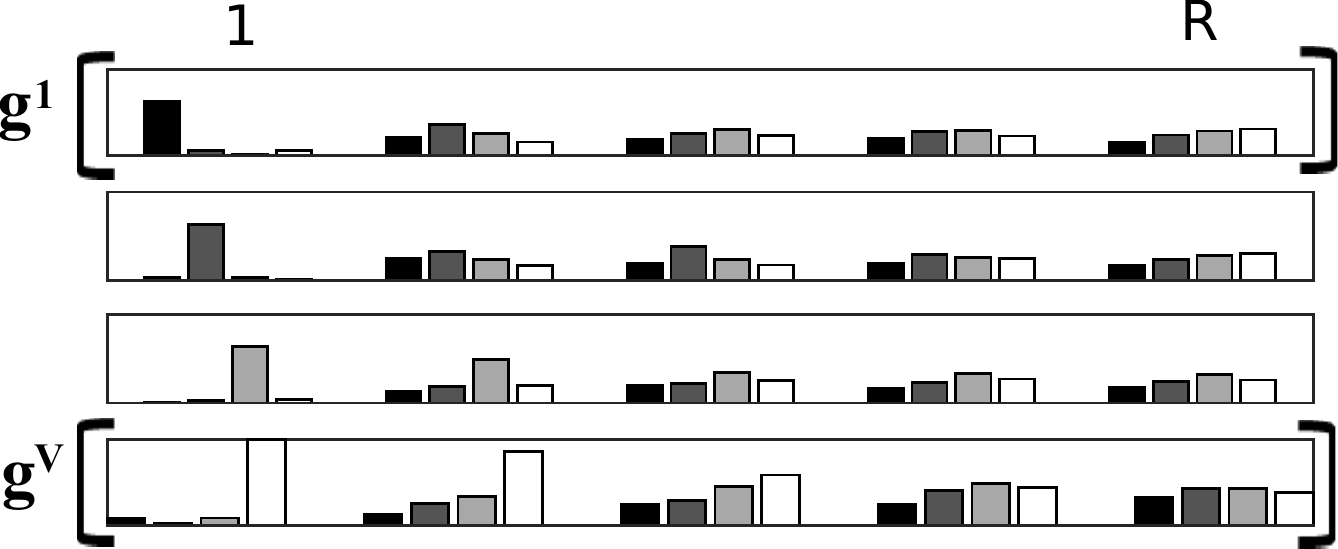}}
\end{center}
\caption{The figure shows the mean vectors $\g^v$ composing feature vector $\f^{ESC}$ corresponding to signs of figure \ref{fig:eSC}.}
\label{fig:HisteSC}
\end{figure}

Histogram $\g^v$ in equation \ref{eq:meanSC} stores the mean values of the shape context of events having direction $v$.
The final feature vector of Event Shape Context concatenates histograms $\g^v$: $\f^{ESC} = [ \g^1 ... \g^V ]$, with $\f^{ESC} \in \Re^{1 \times VRV}$.
Fig. \ref{fig:HisteSC} shows the ESC feature vectors $\f^{ESC}$ of the signs showed in fig. \ref{fig:eSC}.
This methodology characterizes a windows $\w$ of any length $N$ with a more compact representation in a vector of fixed length of $VRV$ features.

To use the eLPB connectivity analysis on this feature, equation \ref{eq:histogramorientedshapecontextevent} is modified and instead of incrementing by one the presence of an event $\e_n$ inside the ring, its eLBP code is considered.
Two families of features can thus be obtained: ESC+eLPB4Pol if the polarity of the events is used to compute the eLBP codes, and ESC+eLBP4Dir if the direction is employed to compute them.
\section{Shape Classification}\label{sec:classification}

This section develops a non-supervised methodology to model each poker symbol of the original Poker-DVS set using HOE and eLBP features. 
Additionally, a second supervised classification framework tackles the 2015 Poker-DVS and the MNIST-DVS recognition using ESC and eLBP.

\subsection{Non-Supervised Generative Gaussian Model}\label{sec:ggm}

Let $\mathcal{E}$ be the set composed of $184,232$ events from the original Poker-DVS recorded data: $\mathcal{E} = \{ \e_1, ... , \e_{184,232} \}$.
The Non-Supervised Generative Gaussian Model seeks to automatically classify the signs using the discriminative power of HOE features.

The analysis of the Poker-DVS dataset is performed by \textit{framing} the set $\mathcal{E}$ into event windows $\w_i = \{ \e_n, ... , \e_{n+N-1} \}$ of $N$ consecutive events.
The descriptors of each $\w_i$ are computed using algorithm \ref{alg:hoe} and algorithm \ref{alg:hoe_elbp}, of section \ref{sec:hoe}, and shown in Fig. \ref{fig:PokerHO2L}.

Dataset $\mathcal{E}$ is split into 3,682 windows $\w_i$ with $N=150$ and an incremental step of $B=50$.
Using this value of $B$, two consecutive $\w_i$ and $\w_{i+1}$ overlap in 100 events.
Events corresponding to the transitions from symbol to symbol are discarded, because the deformation of the shapes can be too severe.
By filtering in this way, the final number of windows is 2,377.
The feature database is then $\mathcal{H}_{N,B} = \{ \f^{HOE}_1, \f^{HOE}_2, ..., \f^{HOE}_{2,363} \}$.

\begin{figure}[h!]
\centerline{
\subfigure[$k=1$ (heart)]{
\begin{tabular}{c}
\includegraphics[width=0.18\columnwidth]{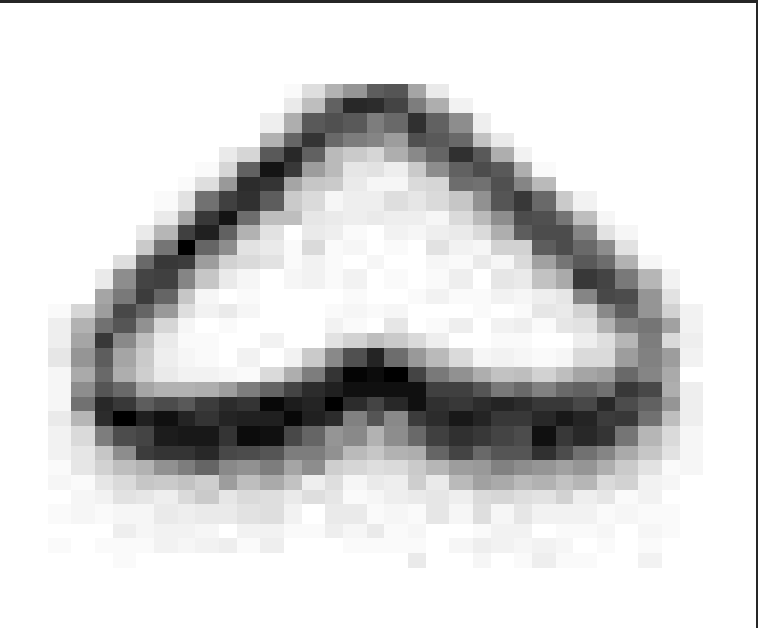} \\
\includegraphics[width=0.22\columnwidth]{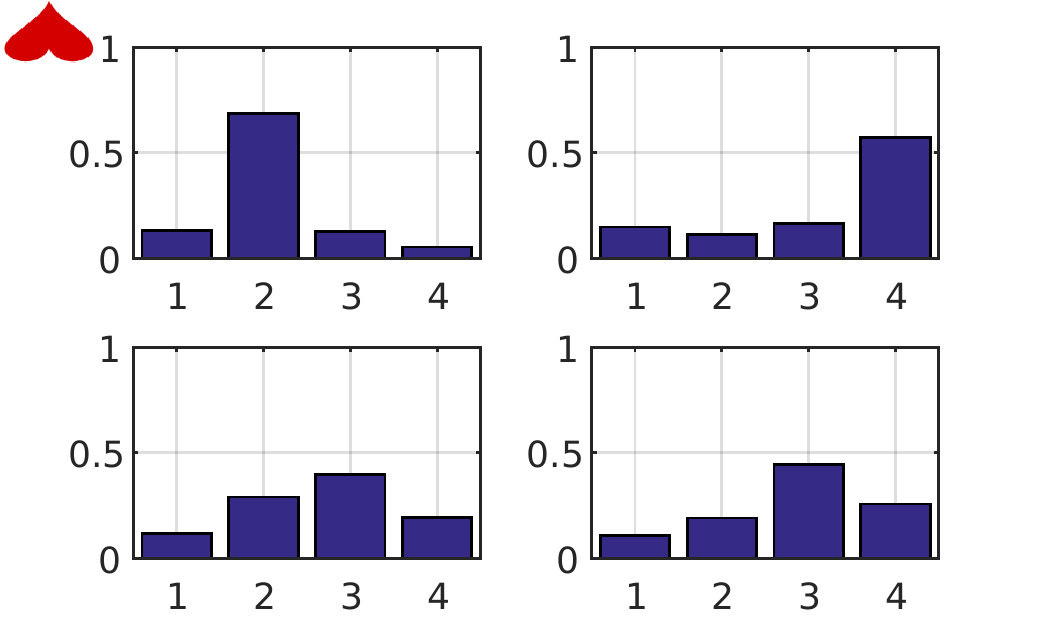} \\
$\m_1$\\
\end{tabular}}
\subfigure[$k=2$ (club)]{
\begin{tabular}{c}
\includegraphics[width=0.18\columnwidth]{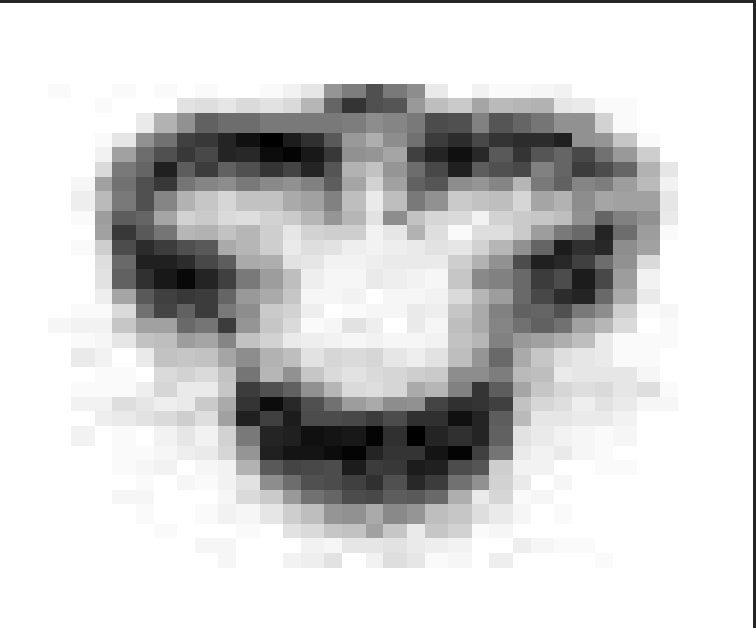} \\
\includegraphics[width=0.22\columnwidth]{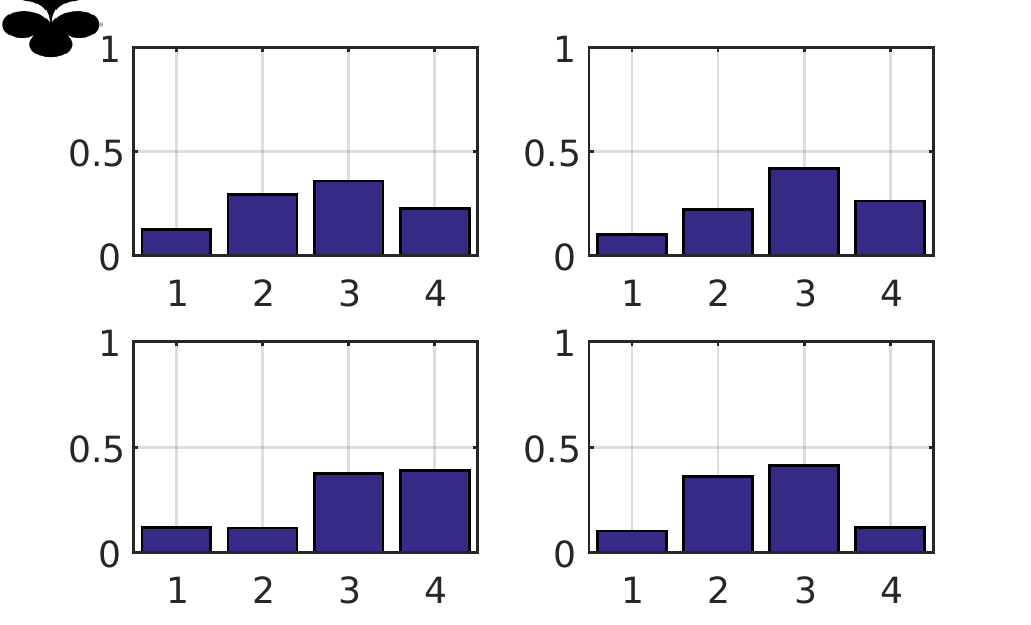} \\
$\m_2$\\
\end{tabular}}
\subfigure[$k=3$ (diamond)]{
\begin{tabular}{c}
\includegraphics[width=0.18\columnwidth]{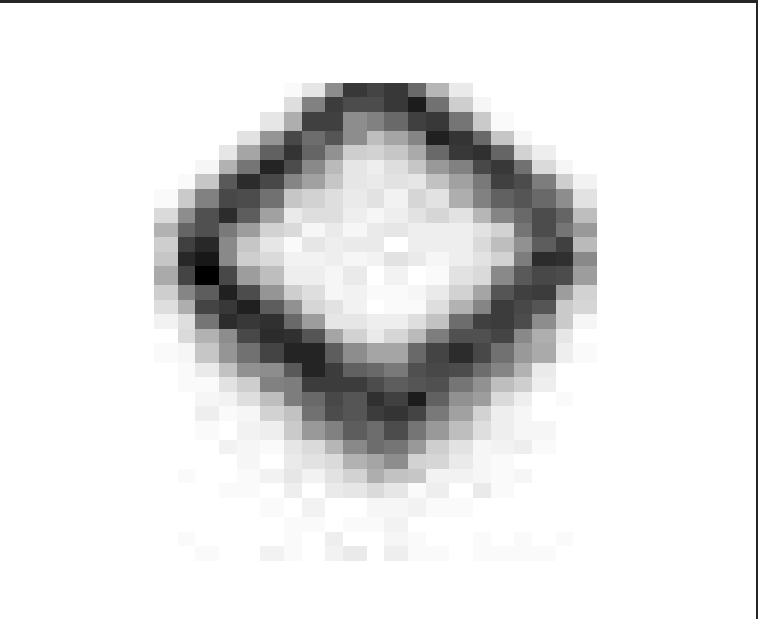} \\
\includegraphics[width=0.22\columnwidth]{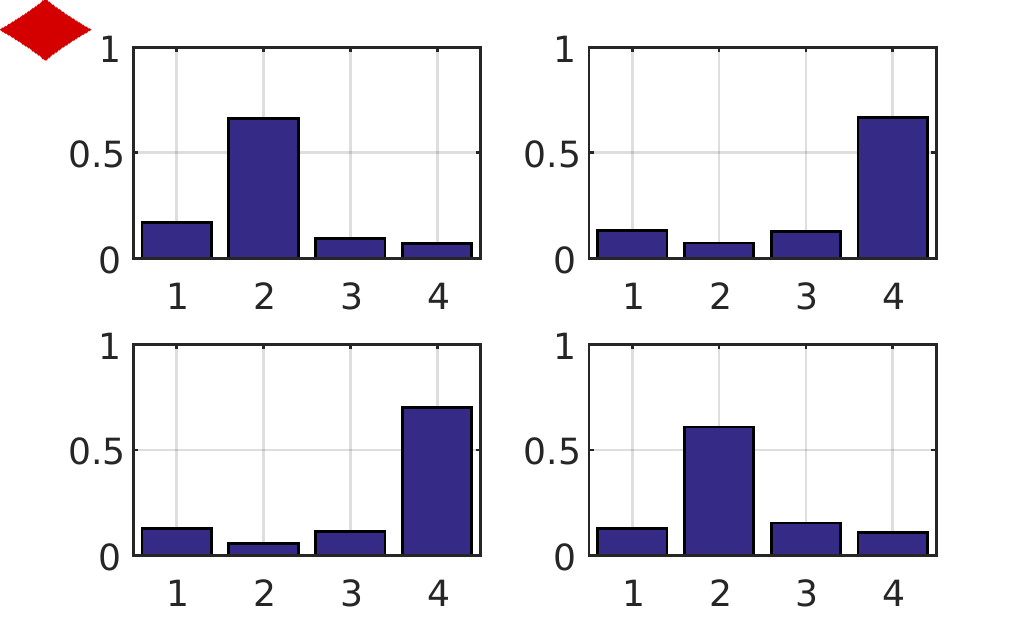} \\
$\m_3$\\
\end{tabular}}
\subfigure[$k=4$ (spade)]{
\begin{tabular}{c}
\includegraphics[width=0.18\columnwidth]{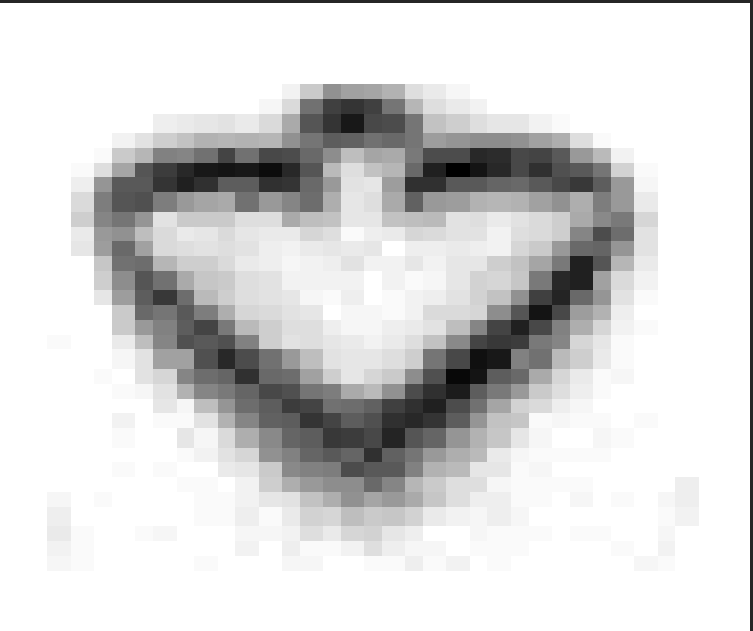} \\
\includegraphics[width=0.22\columnwidth]{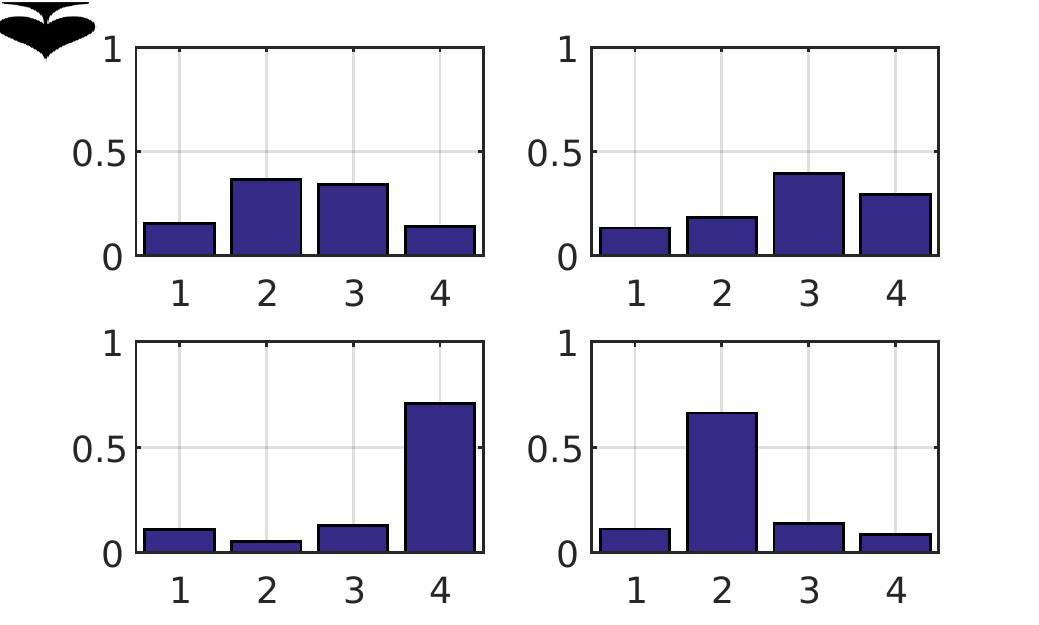} \\
$\m_4$\\
\end{tabular}}}
\caption{Clustering result on Poker Dataset using histogram features, accumulating the $\w_i$ having the same label $k$.}
\label{fig:clustering}
\end{figure}

Each feature vector $\f^{HOE}_i \in \mathcal{H}_{N,B}$ receives a label from a blind K-Means clustering algorithm, with K=4.
Fig. \ref{fig:clustering} shows the result of the K-Means clustering, by histogramming windows $\w_i$ with the same label.
It can be seen that the poker suits are successfully clustered using the feature dataset.
The analysis produced the following result: cluster $k=1$ had a majority of heart sign windows, cluster $k=2$ correspond to club signs, $k=3$ to diamonds, and $k=4$ to spades.

The following step computes a mean histogram $\m_k$ for cluster $k$ using the event windows with label $k$.
Let be $\mathcal{H}_{N,B}^k = \{ \f^{HOE}_1, \f^{HOE}_2, ..., \f^{HOE}_{K_k} \}$, the dataset of cluster $k$ with size $K_k$, and $\f^{HOE}_i$ composed of four histograms: $\f^{HOE}_i=\{\h_{i,1}, \h_{i,2}, \h_{i,3}, \h_{i,4} \}$, corresponding to each quadrant $\kappa_j$.
The mean histogram of $\kappa_j$ is:

\begin{displaymath}
\overline{\h}^k_j = \frac{1}{K_k} \left [ \sum_{n=1}^{K_k} \h_{n,j}(1) \ \ ... \ \ \sum_{n=1}^{K_k} \h_{n,j}(V) \right ]
\end{displaymath}

Each mean histogram $\overline{\h}^k_j$ must be normalized in order to ensure that $\sum_n \overline{\h}^k_j(n)=1$.
Then, the mean histogram of symbol $k$ is defined as: $\m_k =\{\overline{\h}^k_1, \overline{\h}^k_2, \overline{\h}^k_3, \overline{\h}^k_4 \}$.
Fig. \ref{fig:clustering} shows, on the second line, the model histograms of the four poker signs.

The statistical distribution of the Generative Gaussian Model (GGM) is completed by computing a covariance matrix $\S_k$ using a distance function $\d_{i,k}(\f_i,\m_k)$ defined as:

\begin{equation}
\d_{i,k}(\f_i,\m_k)' = \left [ 
\begin{array}{c}
d_e(\h_{i,1},\overline{\h}^k_1) \\
d_e(\h_{i,2},\overline{\h}^k_2) \\
d_e(\h_{i,3},\overline{\h}^k_3) \\
d_e(\h_{i,4},\overline{\h}^k_4) 
\end{array}
\right ]
\end{equation}
\noindent where $d_e(\h_{i,j},\overline{\h}_j)$ is the Euclidean squared distance function between histograms $\h_{i,j}$ and $\overline{\h}_j$.
The cluster $k$ is modeled by duple $(\m_k,\S_k)$.

The classification of event window $\w_i$ in $\mathcal{E}$ using the GGM is defined as:
\begin{equation}
f^{GGM}_k(\w_i) = e^{-( d_{i,k} \S^{-1} d_{i,k})'}
\end{equation}
\noindent being defined as a similarity score employing duple $(\m_k,\S_k)$ to test the feature vector associated to $\w_i$ for cluster $k$.

\subsection{Supervised Multi-Class Recognition}

Fig. \ref{fig:eSC} and Fig. \ref{fig:HisteSC} show the fine difference between \textit{heart} and inverted \textit{spade} shapes.
The non-supervised clustering operation developed in the precedent section did not clearly discriminate them.
A supervised classification approach using Support Vector Machine  (SVM) \cite{Vapnik:1995} is then proposed to enhance those differences and discriminate among the four classes.
Two of the most widely used strategies adapting SVM to multiclass tasks are One-Against-One and One-Against-All \cite{Milgram:2006,Gidudu:2007}.
The One-Against-All approach divides a $K$ class dataset into a $K$ binary SVM classifier.
The One-Against-One approach trains $\left (N \ (N-1) / 2 \right )$ classifiers, each one separating only into two classes.
The output of this kind of framework can be based on the votes that each class received, or by estimating a probability from individual outputs as proposed by Wu \textit{et al.} \cite{Wu:2004}, which is the approach implemented in this paper.
Linear and Radial Basis Function (RBF) kernels are used below to implement the SVM multi-classification.

\subsection{Classification with Memory}

In the testing phase, the probability of input $\w_i$ to belong to sign $k$ is computed as:

\begin{equation}\label{eq:probasWnd}
P_k(\w_i) = \alpha f_k(\w_i) + ( 1 - \alpha ) f_k(\w_{i-1})
\end{equation}
\noindent where $\alpha$ is a memory factor, and $f_k$ is the similarity function of the GMM method or SVM classification function.
Thus $P_k(\w_i)$ uses the current and previous event window classification functions $f_k(\w_i)$ and $f_k(\w_{i-1})$ to smooth the response and become robust to noisy windows.

Sample $\w_i$ is classified as in class $k^*$ which $P^*_k(\w_i)$ produces the largest probability output on eq. \ref{eq:probasWnd}:

\begin{equation}\label{eq:maxWnd}
k^* = argmax_{k=1,2,3,4}P_k(\w_i)
\end{equation}

\section{Experiments and Results}\label{sec:results}

This section implements both classification frameworks on the DVS recorded datasets.

\subsection{Non-supervised GGM Classification Results}

Section \ref{sec:ggm} generated the GGM from event flow $\mathcal{E}$.
This section presents the results of classifying $\mathcal{E}$ with those models to evaluate the discriminative performance of the HOE features.

The GGM classification framework mainly depends on two parameters, the $\alpha$ memory on probability estimation (see eq. \ref{eq:probasWnd}), and the length of the event window $N$.
Further tests were also performed to fix the number of events directions to $V=4$, but are not included in this paper.
For the tests, each event on $\mathcal{E}$ has a label indicating the poker symbol to which it belogns.

Table \ref{tab:featuresResults} presents the sensibility to parameter $\alpha$, using $N=175$ and $B=58$.
Recognition accuracy was obtained computing the mean value of the diagonal of the confusion matrix for the four classes.
Best results were obtained using $\alpha=0.75$ with HOE+eLBP4Pol histogram features framework.
It also outperformed the HOE by 1 \%.
The corresponding confusion matrix is presented on Table \ref{tab:matconf}.

\begin{table}[h!]
\scriptsize
\centering
\begin{tabular}{l|c|c|c|c|c}
\hline
Histogram Feature & $\alpha=1$ & $\alpha=0.75$ & $\alpha=0.5$ & $\alpha=0.25$ & $\alpha=0$\\
\hline
\hline
HOE & 94.56 & 95.35 & 95.25 & 95.10 & 94.91\\
HOE+eLBP4Pol & 94.96 & 96.33 & 95.24 & 96.33 & 95.25\\
HOE+eLPB4Dir & 85.46 & 87.27 & 87.42 & 87.47 & 85.71\\
\hline
\end{tabular}
\caption{Accuracy effect of $\alpha$ parameter.}
\label{tab:featuresResults}
\end{table}

\begin{table}
\scriptsize
\centering
\begin{tabular}{l|c|c|c|c|}
 & \includegraphics[width=0.02\columnwidth]{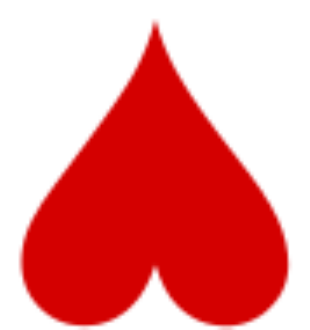} & 
\includegraphics[width=0.02\columnwidth]{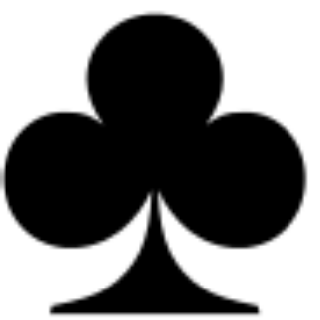}  & 
\includegraphics[width=0.02\columnwidth]{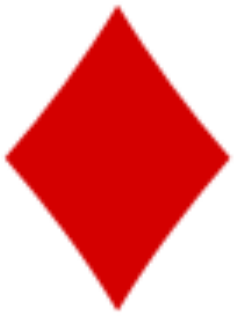} & 
\includegraphics[width=0.02\columnwidth]{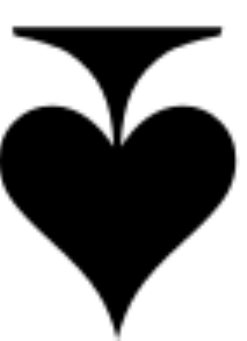} \\
\cline{2-5}
\includegraphics[width=0.02\columnwidth]{pin_heart.pdf} & \textbf{688} & 9 & 2 & 0 \\
\cline{2-5}
\includegraphics[width=0.02\columnwidth]{pin_club.pdf} & 0 & \textbf{447} & 5 & 16 \\
\cline{2-5}
\includegraphics[width=0.02\columnwidth]{pin_diamond.pdf} & 1 & 8 & \textbf{307} & 21 \\
\cline{2-5}
\includegraphics[width=0.02\columnwidth]{pin_spade.pdf} & 0 & 9 & 4 & \textbf{447} \\
\cline{2-5}
\end{tabular}
\caption{The Table shows the Confusion Matrix of the classification using HOE+eLBP4Pol and $\alpha=0.75$.}
\label{tab:matconf}
\end{table}

\begin{figure}[h!]
\centering
\includegraphics[clip,width=0.8\columnwidth]{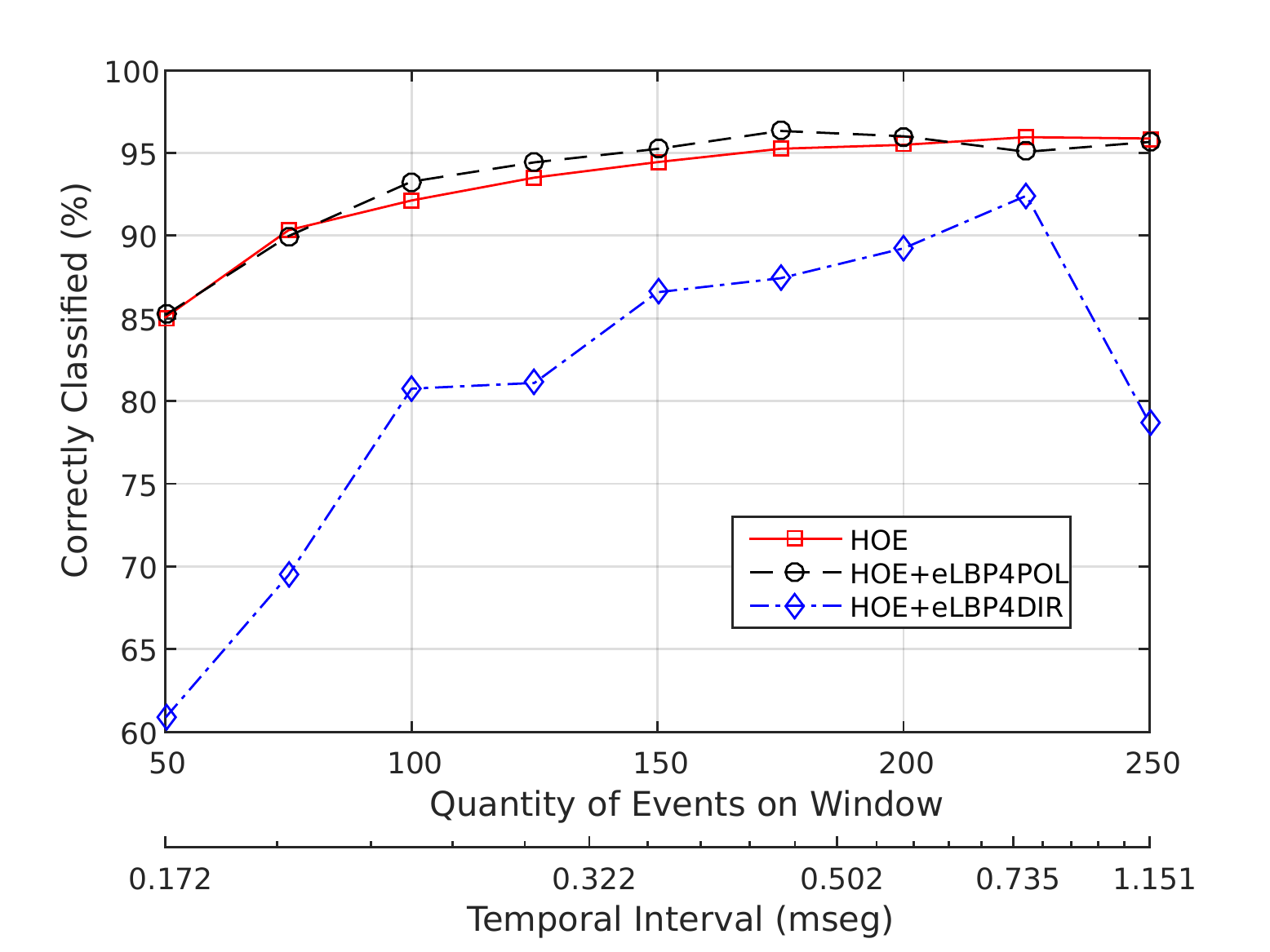} 
\caption{The Figure shows a loglog plot of the Variable Event Length Windows Analysis in terms of the Correct Classifications for the Three Features Histogram Frameworks.}
\label{fig:variablewin}
\end{figure}

Following tests seek to determine how many events are necessary to classify an event window correctly.
Fig. \ref{fig:variablewin} shows the results of the variable windows length $N$ parameter analysis.
The correct classification ratio was maximal at $N=175$ and remained stable for higher $N$.
As a second axis, the mean value of the temporal interval was incorporated  for the corresponding number of events.
Thus, on average, the temporal interval for windows with $N=50$ events was 172 $\mu$seg, and 1151 $\mu$seg for windows of $N=250$.
The HOE+eLBP4Dir features obtained the worst results. 
As shown in the heart example of Fig. \ref{fig:eLBPheart}, this representation gives low weight values to events, which results in a poor characterization of shapes using generative models.

State of the art articles also evaluate the Poker-DVS dataset \cite{Serrano:2015}.
These works employ fixed temporal windows to histogram the events, from 10 ms to 125 ms \cite{Perez:2013}, and 23 ms \cite{Orchard:2015}.
They use supervised classification approaches with complex classifiers: convolutional neural networks (ConNet) \cite{Perez:2013}, spiking neural networks (HFirst) \cite{Orchard:2015}, or events-based time surfaces (HOTS) \cite{Lagorce:2016}.
Table \ref{tab:CGMAccuracy} compares their best results to GGM best performances.

\begin{table}
\centering
\scriptsize
\begin{tabular}{l|c|c|c|c|c}
\hline
Method & N & B & \# windows & Accuracy (\%) & Temporal \\
 &  &  &  &  &  Interval (mseg)\\
\hline
\hline
GGM-HOE & 225 & 75 & 1582 & 96.01 & 0.982\\
GGM-HOE+eLBP4Pol & 175 & 58 & 2044 & 96.33 & 0.640 \\
\hline
ConNet \cite{Perez:2013} & - & - & - & 97.7 & 10-125 \\
HFirst \cite{Orchard:2015} & - & - & - & 97.5 & 23 \\
HOTS \cite{Lagorce:2016} & - & - & - & 95-100 & 20 \\
\hline
\end{tabular}
\caption{Accuracy for the GGM approach and benchmarking with others results of the state of the art.}
\label{tab:CGMAccuracy}
\end{table}

Therefore, in this work, a considerable lower number of events can be employed to obtain equivalent results, in addition to using a very simple classification methodology.

\subsection{Supervised Discriminant Classification}

The supervised discriminant classification methodology was applied to datasets that are more challenging: Poker-DVS and MNIST-DVS \cite{poker_dataset}.

\subsubsection{2015 Poker-DVS dataset}

Tests using a supervised discriminant classification were conducted on the 2015 Poker-DVS dataset \cite{poker_dataset}.
On their website, the authors share a complete recording of the asynchronous events while they were browsing the poker cards, as well as a set of 131 individual files of cropped events.
Each file has a name indicating the sign to which the flow of events corresponds.
A character 'i' is added if the card is inverted.
There are 30 club signs (13 inverted), 43 diamonds (8 inverted), 23 hearts, and 35 spades (10 inverted).
To the best of our knowledge, these are the first results reported on this dataset.

Given the low number of samples per class, the tests were conducted using the Leave-One-Out approach.
This methodology employs all the samples of the set to train the multi-class classifier, except for one sample which is evaluated by the classifier and the result is saved in a confusion matrix.
The overall performance is then obtained by computing the accuracy on the diagonal of the matrix.

The multi-class SVM classifier framework was trained using the LIBSVM library \cite{LIBSVM:2011} and the best parameters for the linear and the RBF kernels were estimated using a 5 cross-fold validation approach.
The framework was composed of the four SVM classifiers, trained using the one-against-one approach.
LIBSVM uses \cite{Wu:2004} to obtain a single probability score for each class $k$: $f_k^{svm}(\w_i)$, with $k=1,2,3,4$.

The framing technique is employed in each test sample to obtain the list of event windows.
Each $\w_i$ is evaluated by the four SVM classifiers using equations \ref{eq:probasWnd} and \ref{eq:maxWnd}, and $\alpha=0.75$ (which gives the best results on the tests).
In this way, the output of the classification accumulates votes for each sign, and the test sample is classified by the sign that receives the highest number of votes.

\begin{figure}[!h]
\centerline{
\subfigure[Linear SVM]{\includegraphics[width=0.49\columnwidth]{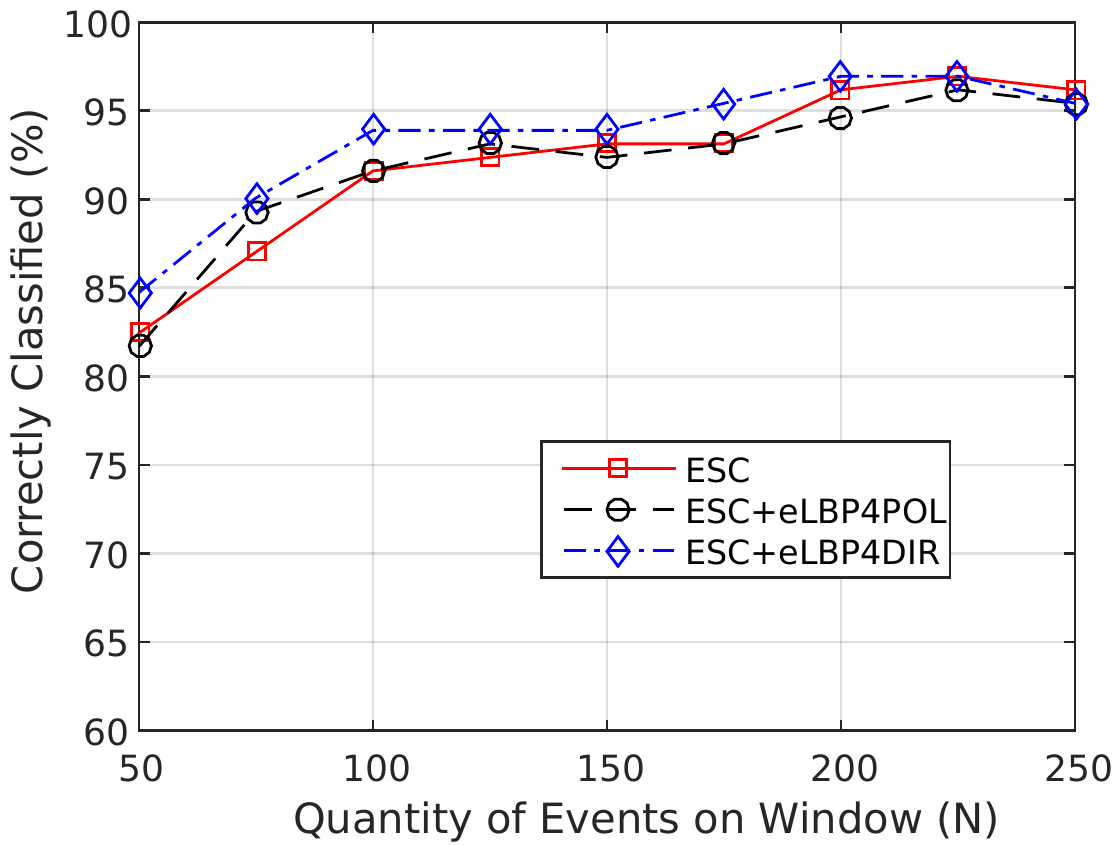}
\label{fig:linsvmres}}
\subfigure[RBF SVM]{\includegraphics[width=0.49\columnwidth]{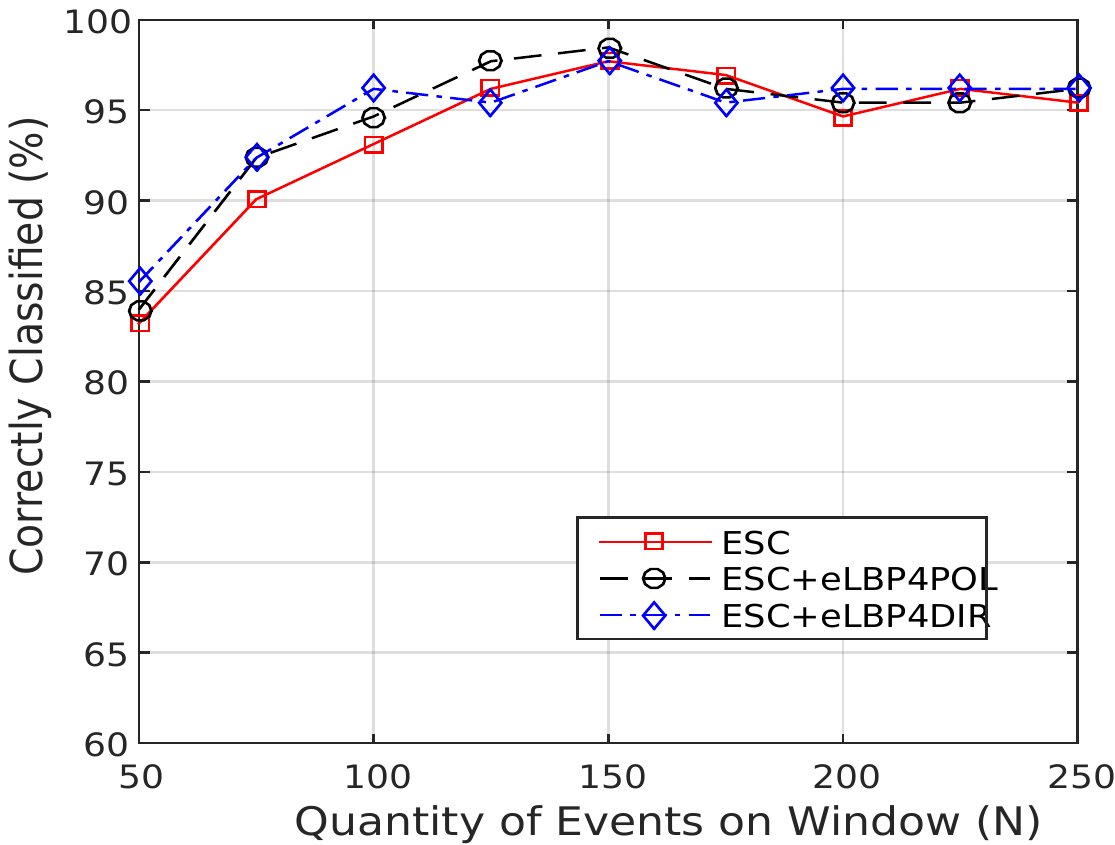}
\label{fig:rbfsvmres}}}
\caption{The figure shows the accuracy of the supervised classification frameworks using Linear or Radial Basis Function kernels on Support Vector Machine classification on events windows of different lengths.}
\label{fig:svmres}
\end{figure}

Fig. \ref{fig:svmres} shows the results obtained for the SVM multi-classification using both kernel functions in different lengths of events windows.

Table \ref{tab:SVMAccuracy} shows the best results, where the total accuracy (TA) is computed adding the diagonal elements of the confusion matrix, divided by the total number of events.
It also presentes the individual accuracy at classification for each windows $\w$A.
The highest score was obtained by the ESC+eLBP4Pol version, yielding a 98.6\% working with a narrow number of events.
It can be observed, that increasing the number $N$ of events improves the individual score $\w$A.

\begin{table}[!h]
\centering
\scriptsize
\begin{tabular}{l|c|c|c|c}
\hline
Descriptor & N & B & TA (\%) & $\w$A  (\%)\\
\hline
\hline
\multicolumn{5}{c}{Linear SVM} \\
\hline
ESC & 225 & 75 & 96.9 &  80.7\\
ESC+eLBP4Pol & 225 & 75 & 96.1 & 81.4 \\
ESC+eLBP4Dir & 225 & 75 & 96.9 & \textbf{81.5} \\
\hline
\multicolumn{5}{c}{RBF SVM} \\
\hline
ESC & 150 & 50 & 97.7 & 75.1\\
ESC+eLBP4Pol & 150 & 50 & \textbf{98.5}  & 76.1 \\
ESC+eLBP4Dir & 150 & 50 & 97.7  & 78.2 \\
\hline
\end{tabular}
\caption{Best results of the SVM classification using the ESC histogram configuration on the 2015 Poker-DVS dataset.}
\label{tab:SVMAccuracy}
\end{table}

The confusion matrix for each best result of the Linear SVM and RBF SVM classifiers is presented on table \ref{tab:svmmatconf}.

\begin{table}[!h]
\scriptsize
\begin{center}
\subfigure[Linear SVM - ESC+eLBP4Pol - $N=225$]{
\begin{tabular}{l|c|c|c|c|}
 & \includegraphics[width=0.02\columnwidth]{pin_heart.pdf} & 
\includegraphics[width=0.02\columnwidth]{pin_club.pdf}  & 
\includegraphics[width=0.02\columnwidth]{pin_diamond.pdf} & 
\includegraphics[width=0.02\columnwidth]{pin_spade.pdf} \\
\cline{2-5}
\includegraphics[width=0.02\columnwidth]{pin_heart.pdf} & \textbf{22} & 1 & 0 & 0 \\
\cline{2-5}
\includegraphics[width=0.02\columnwidth]{pin_club.pdf} & 0 & \textbf{28} & 1 & 1 \\
\cline{2-5}
\includegraphics[width=0.02\columnwidth]{pin_diamond.pdf} & 0 & 1 & \textbf{42} & 0 \\
\cline{2-5}
\includegraphics[width=0.02\columnwidth]{pin_spade.pdf} & 0 & 0 & 0 & \textbf{35} \\
\cline{2-5}
\end{tabular}
\label{tab:linconfmat}}
\subfigure[RBF SVM - ESC+eLBP4Pol - $N=150$]{
\begin{tabular}{l|c|c|c|c|}
 & \includegraphics[width=0.02\columnwidth]{pin_heart.pdf} & 
\includegraphics[width=0.02\columnwidth]{pin_club.pdf}  & 
\includegraphics[width=0.02\columnwidth]{pin_diamond.pdf} & 
\includegraphics[width=0.02\columnwidth]{pin_spade.pdf} \\
\cline{2-5}
\includegraphics[width=0.02\columnwidth]{pin_heart.pdf} & \textbf{22} & 1 & 0 & 0 \\
\cline{2-5}
\includegraphics[width=0.02\columnwidth]{pin_club.pdf} & 0 & \textbf{29} & 0 & 1 \\
\cline{2-5}
\includegraphics[width=0.02\columnwidth]{pin_diamond.pdf} & 0 & 0 & \textbf{43} & 0 \\
\cline{2-5}
\includegraphics[width=0.02\columnwidth]{pin_spade.pdf} & 0 & 0 & 0 & \textbf{35} \\
\cline{2-5}
\end{tabular}
\label{tab:rbfconfmat}}
\end{center}
\caption{Confusion Matrix of the best results for Linear SVM and RBF SVM classifiers.}
\label{tab:svmmatconf}
\end{table}

Linear SVM results in Fig. \ref{fig:linsvmres} show a that ESC and ESC+eLBP4Pol got similar scores, and ESC features weighted by the eLBP using directions performed better for low values of $N$.
Performance of the ESC feature was very satisfactory showing that it is robust enough to the inversion of the signs.
There may be few confused samples showing that the inversion is well conducted by the Linear classifier and the ESC features.

On the other hand, SVM classifier using RBF kernel functions increased the discriminant power of the ESC feature weighted by the eLBP using polarity, which got the highest accuracy.
This proves that this kind of connectivity analysis of events characterizes the boundaries of the shape successfully.
As can be seen on the confusion matrix of table \ref{tab:svmmatconf}, there are only two mistakes: one associated with sign inversion (club) and the other one related to the very low number of events (heart) in the sequence.

\subsubsection{2015 MNIST-DVS dataset}

\begin{figure}[!h]
\centerline{
\includegraphics[width=0.6\columnwidth]{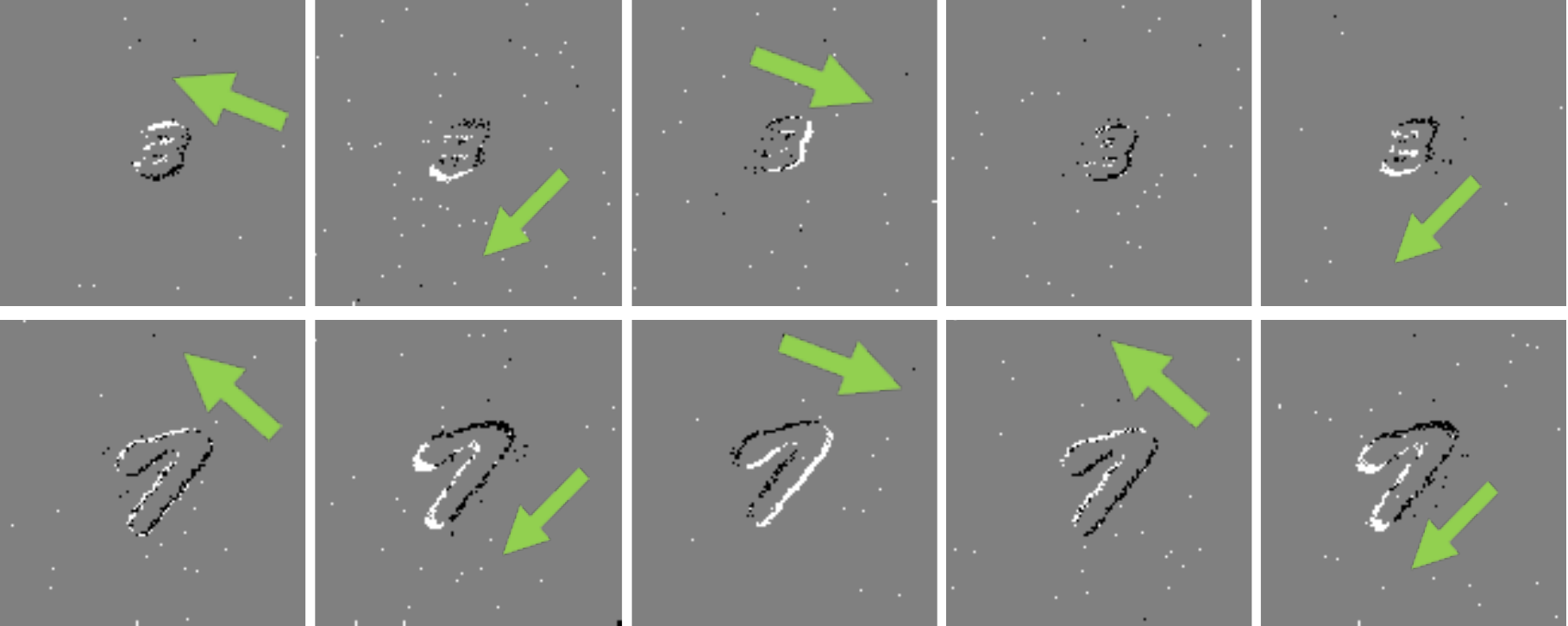}}
\caption{MNIST-DVS digits samples at the three available scales.}
\label{fig:mnist-dvs}
\end{figure}

MNIST-DVS dataset \cite{Serrano:2015} was obtained by transforming the original MNIST digits dataset \cite{LeCun:1998} into an event-driven stream.
Each digit was visualized on a screen and captured by the DVS camera while moving slowly and following a random path.
The frame-based data was thus transformed into an asynchronous-event based representation.
By using this protocol, 10,000 digits were captured at 3 different scales: 4, 8 and 16.
Fig. \ref{fig:mnist-dvs} shows examples of 4 and 8 scales.
The number $N$ of events within an event window $\w_N$ was empirically fixed to $N=300$ and $N=600$, respectively.
In Fig.~\ref{fig:mnist-dvs}, rows show the direction of the motion of the corresponding $\w_N$.
It is possible to verify the effect of invisible edges that are on the direction of the movement (see Fig. \ref{fig:featuresGrids}).
Raw files were employed for the tests, instead of using script provided on the web site to filter and centralize the event streams, because this script changes the time stamps information which is important for the implementation of the ESC features.

ESC feature extraction methodology was implemented by using the rings configuration as detailed on Sec. \ref{sec:ESC}, along with the same value for the variables: $R=5$ and $V=4$.
A configuration for ESC where each ring is split into two cells or hemispheres, was also proposed, differentiating events placed above or below of the studied event, considering the vertical axis.
Therefore, the feature captured information about the vertical orientation of the shape, and helped to discriminate digit ``6'' from digit ``9''.
The size of the feature vector for each $\w_N$ was then $2VRV=160$ elements.
A 10-fold cross validation is employed for training the supervised SVM classification using a RBF kernel.
Global performance was computed as the average value of the accuracy in each fold.

Table~\ref{tab:MNISTAccuracy} depicts the results of supervised classification system using both representations of the ESC on the smallest scale 4.
TA is the accuracy of the complete stream of events corresponding to one digit, and $\w$A is the accuracy of individuals windows.
Results were compared to those of Zaho \textit{et al.} \cite{Zhao:2015} and Hederson \textit{et al.} \cite{Henderson:2015}, which are reported in \cite{Serrano:2015}.
The results for ESC feature representation considering one cell per ring are $5RC1$.
The representation with two cells by ring is denominated $5RC2$.

\begin{table}
\centering
\scriptsize
\begin{tabular}{|l|c|c|c|c|c|c|}
\hline
  & \multicolumn{3}{|c|}{scale 4 - $N = 300$}  & \multicolumn{3}{|c|}{scale 8 - $N = 600$}\\
\hline
Method &  TA & $\w$A & Tmp &  TA & $\w$A & Tmp  \\
 &    (\%) & (\%) &  (ms) &    (\%) & (\%) &  (ms) \\
\hline
\multicolumn{7}{|l|}{5RC1}\\
\hline
RBF-ESC & 88.42 & 74.05 &   & 90.6 & 79.10 &   \\
RBF-ESC+eLBP4Pol & 88.02 & 73.02 & 95 & 90.6 & 78.55 & 87 \\
RBF-ESC+eLBP4Dir & 88.15 & 73.29 &    & 90.0 & 77.99 &    \\
\hline
\multicolumn{7}{|l|}{5RC2}\\
\hline
RBF-ESC & \textbf{94.6} & 84.78 &    & \textbf{96.3} & 90.47 &    \\
RBF-ESC+eLBP4Pol & 94.3 & 84.37 & 95 & 96.0 & 89.81 & 87  \\
RBF-ESC+eLBP4Dir & 93.9 & 84.08 &    & 96.1 & 89.86 &    \\
\hline
\hline
Zhao \cite{Zhao:2015}  & 78.86 & - & 100 & & &\\
Zhao \cite{Zhao:2015}  & 88.14 & - & 2000 & & &\\
Henderson \cite{Henderson:2015} & 87.41 & - & 2000 & & &\\
\hline
\end{tabular}
\caption{Accuracy of the supervised approach on the MNIST-DVS dataset and benchmarking with others results of the state of the art.}
\label{tab:MNISTAccuracy}
\end{table}

The representation of ESC that incorporated the vertical orientation (5RC2) of the digits yielded the best results: 94.6 \% and 96.3 \%, for scales 4 and 8 respectively.
Additionally, scale 4 results outperformed the best of Zhao and Henderson by 6 \%, comparing the cases where the complete stream of a digit had been evaluated.
Zhao's thest using temporal windows of 100 ms is comparable to the results of $\w$A column which have a temporal length of 95 ms on average. and and improvement in accuracy of about 6 \%.

As can be seen in Table \ref{tab:MNISTAccuracy}, the use of the connectivity operator did not improve the results with these values of $N$.
Connectivity analysis can be useful for lower values of $N$, as shown on the Poker-DVS dataset enhancing the histogram representation.
Its effect is relative when using high values of $N$.
In addition, both HOE and ESC features are sensible to higher values of $N$, as highlighted in Fig. \ref{fig:svmres} and Fig. \ref{fig:variablewin}.
Considerably increasing $N$, increases the temporal window and the shape suffers a deformation, as shown in Fig. \ref{fig:disco_experiment}. 
Local features lose then their characterization power.

\section{Conclusions and Discussion} \label{sec:conclusions}

Traditional histogram representations were evaluated to exploit event-based information provided by DVS.
They were carefully adapted to a spatio-temporal representation, switching the accumulators behavior to a 3D representation.
A simple operator was also proposed to measure events connectivity, referred to as eLBP Codes.
This original feature extraction methodology enriched information provided by DVS cameras technologies.
It was found that histograms representation that employ events orientations have a good performance characterizing non-deformable shapes, outperforming state-of-art results.

Further research should include the characterization of deformable objects such as human silhouettes and faces.
To this end, histograms features could be computed using a multi-scales temporal approaches.
For instance, the Hierarchy of Time Surfaces (HOTS) approach \cite{Lagorce:2016}, incorporates iteratively larger spatial and temporal spaces, i.e. Time-Surfaces, where specific descriptors are computed and delivered to the next scale or layer. 
This feature extraction process could be implemented using HOE and ESC to detect deformable objects at different scales.

\section*{Acknowledgments}

This paper was supported by PID P16T01 of UADE.
The author thanks Dr. Bernab\'{e} Linares-Barranco for providing the original Poker-DVS dataset.

\bibliographystyle{elsarticle-num}

\end{document}